\documentclass{scaleai-paper}
\usepackage[square,sort&compress,numbers]{natbib}

\usepackage{mathpazo}

\usepackage[scaled=0.92]{helvet}
\usepackage{fontawesome5}
\usepackage{subcaption}

\usepackage{hyperref}       
\hypersetup{
    colorlinks=true,
	linkcolor=blue,
	filecolor=magenta,
	urlcolor=blue,
	citecolor=black,
	pdfinfo={
        Title={LHAW: Controllable Underspecification for Long-Horizon Tasks},
        Keywords={Agents, Benchmarks, Clarification, Human-in-the-Loop, Long-Horizon Tasks, Underspecification, Synthetic Data},
    }
}
\usepackage{longtable}
\usepackage{tabularx}
\usepackage[utf8]{inputenc} 
\usepackage[T1]{fontenc}    
\usepackage{url}            
\usepackage{booktabs}       
\usepackage{amsfonts}       
\usepackage{nicefrac}       
\usepackage{microtype}      
\usepackage{multirow}
\usepackage{amsmath}
\usepackage{amssymb}
\usepackage{mathtools}
\usepackage{amsthm}
\usepackage{xspace}
\usepackage{graphicx}
\usepackage[table]{xcolor}
\usepackage{array}
\usepackage{colortbl}
\usepackage{placeins}

\usepackage{enumitem}
\setlist[itemize]{leftmargin=*}
\setlist[enumerate]{leftmargin=*}
\setlist{noitemsep,topsep=2pt,parsep=2pt,partopsep=0pt}

\usepackage{tcolorbox}
\usepackage{soul}
\usepackage{cleveref}
\usepackage{listings}
\usepackage{tikz}
\usepackage{eso-pic}

\theoremstyle{plain}
\newtheorem{theorem}{Theorem}[section]

\theoremstyle{definition}
\newtheorem{definition}[theorem]{Definition}

\theoremstyle{remark}

\let\svthefootnote\thefootnote
\newcommand\freefootnote[1]{%
  \let\thefootnote\relax%
  \footnotetext{#1}%
  \let\thefootnote\svthefootnote%
}

\makeatletter
\renewcommand\AB@affilsepx{, \protect\Affilfont}
\makeatother

\newcommand\blfootnote[1]{
    \begingroup
    \renewcommand\thefootnote{}\footnote{#1}
    \addtocounter{footnote}{-1}
    \endgroup
}

\newcommand{\passmark}{\texttt{pass@k}}
\newcommand{\lhaw}{\textsc{LHAW}}
\newcommand{\tac}{\textsc{TheAgentCompany}}
\newcommand{\mcpatlas}{\textsc{MCP-Atlas}}
\newcommand{\swebench}{\textsc{SWE-Bench Pro}}

\newcommand{\deltaval}[1]{{\color{gray}\footnotesize~(#1)}}

\title{LHAW: Controllable Underspecification for Long-Horizon Tasks}

\author{George Pu$^*$}
\author{Michael S. Lee$^*$}
\author{Udari Madhushani Sehwag}
\author{David J. Lee}
\author{\authorcr Bryan Zhu}
\author{Yash Maurya}
\author{Mohit Raghavendra}
\author{Yuan Xue}
\author{and Samuel Marc Denton}

\affil{Scale AI}

\newcommand{\authoremail}{%
  \vspace{-1.5em}
  \faEnvelope\ \blackmailto{george.pu@scale.com}
  \quad
  \faDatabase\
  \href{https://huggingface.co/datasets/ScaleAI/lhaw}{huggingface.co/ScaleAI/lhaw}
  \quad
  \faGlobe\
  \href{https://github.com/scaleapi/lhaw/}{github.com/scaleapi/lhaw/}
  \vspace{-1em}
}

\begin{document}

\newcommand*\circled[1]{\tikz[baseline=(char.base)]{
            \node[shape=circle,draw,inner sep=1pt] (char) {#1};}}
\newcommand{\blackmailto}[1]{%
  {\hypersetup{urlcolor=black}\href{mailto:#1}{\texttt{#1}}}%
}

\maketitle
\authoremail

\blfootnote{$^*$Indicates equal contribution}
\begin{abstract}
Long-horizon workflow agents that operate effectively over extended periods are essential for truly autonomous systems.
Their reliable execution critically depends on the ability to reason through ambiguous situations in which clarification seeking is necessary to ensure correct task execution.
However, progress is limited by the lack of scalable, task-agnostic frameworks for systematically curating and measuring the impact of ambiguity across custom workflows. 
We address this gap by introducing \lhaw{} (\textbf{L}ong-\textbf{H}orizon \textbf{A}ugmented \textbf{W}orkflows), a modular, dataset-agnostic synthetic pipeline that transforms any well-specified task into controllable underspecified variants by systematically removing information across four dimensions -- Goals, Constraints, Inputs, and Context -- at configurable severity levels.
Unlike approaches that rely on LLM predictions of ambiguity, \lhaw{} validates variants through empirical agent trials, classifying them as outcome-critical, divergent, or benign based on observed terminal state divergence.
We release 285 task variants from \tac{}, \swebench{} and \mcpatlas{} according to our taxonomy alongside formal analysis measuring how current agents detect, reason about, and resolve underspecification across ambiguous settings.
\lhaw{} provides the first systematic framework for cost-sensitive evaluation of agent clarification behavior in long-horizon settings, enabling development of reliable autonomous systems.
\end{abstract}

\section{Introduction}
\label{sec:introduction}

Autonomous agents executing complex, multi-step workflows over extended time horizons face a fundamental challenge: distinguishing between tasks that are solvable given current information and those that require clarification to avoid execution failures. Unlike conversational assistants where clarification incurs low cost, autonomous agents operate in situations where each human interaction interrupts workflow execution, introduces latency, and imposes cognitive switching costs on supervisors. The critical question becomes not whether agents can execute well-specified tasks, but if they can detect when specifications are insufficient and strategically decide when clarification justifies its cost.

Current benchmarks evaluate agents along dimensions that fail to capture this tension. Long-horizon benchmarks like \tac{} \cite{theagentcompany2024}, \swebench{} \cite{swebench2024} and \mcpatlas{} \cite{bandi2026mcpatlaslargescalebenchmarktooluse} assess execution capability under sufficient specification without assessing whether agents recognize missing information. Clarification benchmarks like ClariQ \cite{aliannejadi2020convai3} and AmbigQA \cite{min2020ambigqa} operate in short-context regimes where asking questions carries negligible cost. VitaBench \cite{he2025vitabench} explores these types of complex interactions with users, but does not systematically evaluate whether agents can detect underspecification in long-horizon tasks and appropriately seek clarification when needed. An agent that silently assumes missing parameters may execute incorrect workflows with cascading consequences, whereas an agent that asks excessively may indicate poor reasoning about what information gaps are critical. The absence of systematic evaluation for this capability impedes progress toward reliable autonomous systems.

We introduce \lhaw{}, a dataset-agnostic pipeline for generating and validating underspecified task variants to address this gap. \lhaw{} transforms well-specified benchmark tasks into controllably underspecified variants through three phases: (1) \textbf{Segment Extraction}---identifying removable information segments categorized by dimension (Goal, Constraint, Input, Context) and scored for criticality; (2) \textbf{Candidate Generation}---producing underspecified variants at configurable severity levels; and (3) \textbf{Empirical Agent Trials}---classifying each variant as \emph{outcome-critical} (removal causes consistent failure), \emph{divergent} (variable outcomes), or \emph{benign} (agents infer the missing information) based on observed agent behavior. This operationalizes ambiguity through execution rather than linguistic intuition, grounding evaluation in observable outcomes and allowing us to make guarantees about the level of ambiguity. 

Key contributions of our work are:

\begin{enumerate}
\item \textbf{A synthetic pipeline for controllable underspecification} (\S\ref{sec:benchmark}-\S\ref{sec:pipeline}). \lhaw{} transforms well-specified tasks into underspecified variants via three phases: segment extraction, candidate generation, and empirical validation. The pipeline operates over four information dimensions (Goal, Constraint, Input, Context) with configurable obfuscation strategies.

\item \textbf{Empirical validation and benchmark-ready samples} 
(\S\ref{sec:benchmark-ready}-\S\ref{sec:pipeline-validation}). We generate 285 underspecified variants across \tac{}, \swebench{}, and \mcpatlas{}, empirically classifying each as outcome-critical, divergent, or benign through agent trials. Severity strategies and multi-segment removal produce predictable difficulty curves, validating controllable ambiguity levels. 

\item \textbf{Analysis of clarification behavior under underspecification} (\S\ref{sec:experiments}). Frontier models consistently recover significant performance through clarification, but with clear model-dependent patterns: GPT-5.2 over-clarifies while Gemini models under-clarify. We introduce Gain/Q (performance gain per question) to measure clarification efficiency, revealing which models extract the most value per question—and which impose unnecessary burden on users. 

\end{enumerate}

\begin{figure*}[!t]
\begin{center}
\centerline{\includegraphics[width=0.95\textwidth]{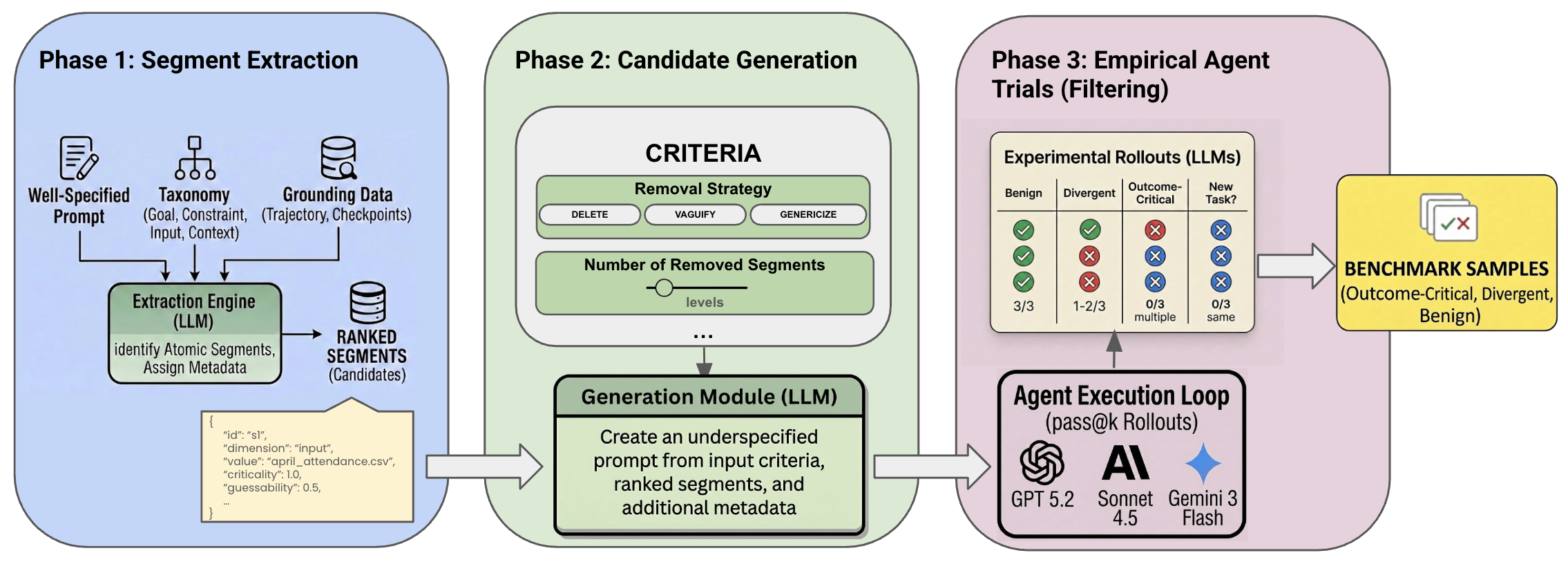}}
\caption{\textbf{The \lhaw{} Synthetic Underspecification Pipeline.} Three phases: (1)~\emph{Segment Extraction \& Scoring}---identify removable segments, classify by dimension, estimate criticality/guessability; (2)~\emph{Variant Generation}---apply different removal strategies to create underspecified prompts with expected questions; (3)~\emph{Empirical Validation}---run agent trials to classify variants as outcome-critical, divergent, or benign.}
\label{fig:pipeline}
\end{center}
\vskip -0.1in
\end{figure*}

\section{Background and Problem Setup}
\label{sec:background}

\subsection{Long-Horizon Workflows}

Traditional AI assistants operate in a \emph{conversational} paradigm: rapid turn-by-turn dialogue where clarification is cheap and expected. In some scenarios, engagement is a primary success metric. However, the emerging paradigm of \emph{long-horizon workflows} inverts this: agents execute extended workflows with minimal supervision, and each human interruption carries a cost.

\begin{definition}[Long-Horizon Workflow]
A task requiring an agent to execute dozens to hundreds of sequential actions across multiple stages, where human interaction is available but costly.
\end{definition}

Clarification carries costs along three dimensions: (1)~\emph{interruption}---human context-switching and cognitive load; (2)~\emph{latency}---workflow blocked pending response; and (3)~\emph{trust}---excessive questions erode user confidence. An ideal agent clarifies when the expected value of information exceeds these costs, and proceeds autonomously otherwise. Toward this goal, we study the value of information and the cost of corresponding questions separately, with their integration into a decision-theoretic framework left for future work.

\subsection{Two Types of Ambiguity}\label{sec:ambig_types}

Scenarios that warrant clarification can be broadly categorized into two types:

\begin{enumerate}
    \item \textbf{Semantic Ambiguity.} A request admits multiple valid interpretations, each leading to a different but potentially complete answer. This is common in open-domain QA and preference elicitation; benchmarks like ClariQ \cite{aliannejadi2020convai3} and AmbigQA \cite{min2020ambigqa} evaluate these subjective scenarios.
    \item \textbf{Underspecification.} A request is missing critical information, rendering it unsolvable without acquiring that information. This is frequent in enterprise settings where tasks are complex and users may not know what details are necessary. Benchmarks like QuestBench \cite{questbench2024} and UserBench \cite{userbench2024} explore this setting. 
\end{enumerate} 

\begin{definition}[Outcome-Critical Underspecification]
A task is outcome-critical if missing information causes consistent agent failure, showing that the agent cannot succeed without acquiring the underspecified information.
\end{definition}

\lhaw{} focuses on \emph{underspecification}---cases where agents cannot succeed without acquiring missing information. We validate variants through empirical agent trials, classifying them as outcome-critical, divergent, or benign based on observed terminal state divergence and agent behaviors.

\section{The \lhaw{} Pipeline}
\label{sec:benchmark}

\lhaw{} is a synthetic pipeline for controllable underspecification, designed around long-horizon tasks and generality. We step through the process of creating our benchmark-ready samples by first introducing task selection (\S\ref{sec:tasks}), then introducing our underspecification taxonomy and synthetic pipeline (\S\ref{sec:taxonomy}-\S\ref{sec:pipeline}), and wrap up with the benchmark, user-simulator design and pipeline validation (\S\ref{sec:benchmark-ready}-\S\ref{sec:pipeline-validation}).

\subsection{Task Selection and Evaluation Design}
\label{sec:tasks}

\lhaw{} evaluates strategic clarification on long-horizon tasks drawn from three benchmarks spanning tool-use and real-world workflows: \tac{} (OwnCloud subset), \swebench{}, and \mcpatlas{}. Our goal is to study underspecification in settings where agents can make meaningful progress when all information is present, but may fail when critical details are removed.

\paragraph{Selection Criteria.}
To reduce confounding factors in agent error traces, we select tasks where agents demonstrate \emph{baseline competence} under fully specified prompts, then systematically ablate outcome-critical information. This ensures that performance drops under underspecification reflect ambiguity sensitivity rather than general task inability.

\paragraph{Evaluation Metrics.}
We report \textbf{pass@3} as full task success: at least one of three independent trajectories satisfies all evaluation criteria. When full success is too sparse to support filtering, we use \textbf{checkpoint-level progress} to identify tasks where agents reach ambiguity-sensitive decision points.

\paragraph{Partial Completion Tasks (TAC).}
\tac{}~\cite{theagentcompany2024} is a realistic, workplace-oriented benchmark in which agents act as digital workers within a simulated software company. Tasks require navigation of multi-step workflows involving web browsing, code editing, terminal commands, and collaboration tools, spanning Data Science, Finance, HR, and Software Engineering domains.

Each task defines multiple \emph{checkpoints} corresponding to meaningful sub-goals. Full end-to-end success is rare---only 7 of 33 OwnCloud tasks achieve non-zero pass@3 across frontier models---making filtering by full completion impractical. Instead, we select by \emph{partial completion}: tasks with average checkpoint accuracy $\geq$50\% across reference models (Sonnet-4.5, GPT-5.2, Gemini-3-Flash), which yields 13 tasks (Table~\ref{tab:tasks}).

\paragraph{Full Completion Tasks (SWE-Bench-Pro, MCP-Atlas).}
\swebench{}~\cite{swebench2024} evaluates code repair capabilities on real GitHub issues, requiring agents to navigate large codebases and produce correct patches. For this benchmark with higher pass rates, we require pass@3 $> 0$ for \emph{all} reference models (GPT-5.2, Sonnet-4.5, Gemini-3-Pro)---ensuring tasks are solvable end-to-end before underspecification is introduced. For \swebench{}, each task includes \emph{fail-to-pass} (F2P) tests---unit tests that must change from failing to passing for a correct fix---and \emph{pass-to-pass} (P2P) regression tests that must remain passing. We use F2P test outcomes as checkpoints for partial progress and behavioral divergence classification; P2P tests are not used for checkpoints but are required for full task success (no regressions). To enable meaningful divergence analysis, we require $|$F2P$|$ $> 2$ (at least 3 tests), reducing 731 instances to 417. Filtering further by pass@3 $> 0$ for all reference models yields 100 base tasks. From these, we generate underspecified variants and classify each via empirical trials (\S\ref{sec:classification}), selecting 100 variants spanning 75 unique original tasks with a representative distribution of 31 outcome-critical, 34 divergent, and 35 benign.

\mcpatlas{} \cite{bandi2026mcpatlaslargescalebenchmarktooluse} tests tool-use capabilities across diverse MCP server integrations, requiring agents to select and invoke appropriate tools for each task. For this benchmark, we similarly require pass@3 $> 0$ for all reference models (GPT-5.2, Opus-4.5, Gemini-3-Pro) where a pass is defined as achieving $>=$ 75\% of checkpoints -- the default threshold in the official github repository. This yields 222 tasks, of which we generate 666 variants (3 per task) and then randomly downsample to 100 variants to reach our desired distribution of 50 outcome-critical, 30 divergent, and 20 benign.

\subsection{Underspecification Taxonomy}
\label{sec:taxonomy}

\begin{table*}[t]
\caption{Underspecification taxonomy: four dimensions of task information, with definitions, indicators, and examples.}
\label{tab:taxonomy}
\vskip 0.1in
\begin{center}
\begin{small}
\begin{tabular}{@{}p{1.4cm}p{4.2cm}p{3.8cm}p{5.5cm}@{}}
\toprule
\textbf{Dimension} & \textbf{Definition} & \textbf{Indicators} & \textbf{Examples} \\
\midrule
\textbf{Goal} & The final deliverable or success criteria is unclear. Agent doesn't know what output to produce or what counts as completion. & Vague action verbs, missing deliverable spec, undefined success criteria, ambiguous scope & ``Write the summary.'' $\rightarrow$ What summary? For whom? What length? \newline ``Launch the campaign.'' $\rightarrow$ What is the success metric? \\
\addlinespace
\textbf{Constraint} & Rules, thresholds, methods, or requirements for execution are unclear. Agent knows \emph{what} but not the specific parameters. & Missing thresholds, filters, limits, timelines, exclusion criteria, vague precision & ``List top customers.'' $\rightarrow$ What defines ``top''? By revenue? Count? \newline ``Book a reasonable flight.'' $\rightarrow$ What's your budget? \\
\addlinespace
\textbf{Input} & Data source, file location, system, or tool to use is unclear. Agent cannot locate or access required resources. & Unspecified data source, file, system, version; ambiguous ``latest'' qualifier & ``Use the latest dataset.'' $\rightarrow$ Which dataset? Stored where? \newline ``Run analysis on the customer table.'' $\rightarrow$ Which database? \\
\addlinespace
\textbf{Context} & Domain knowledge, business logic, jargon, or implicit requirements are unclear. Agent lacks understanding to interpret correctly. & Missing actor/recipient, referent ambiguity, domain or tool context & ``Send the report to the team.'' $\rightarrow$ Which team? Email or chat? \newline ``Deploy after tests.'' $\rightarrow$ Which tests? What tools? \\
\bottomrule
\end{tabular}
\end{small}
\end{center}
\vskip -0.1in
\end{table*}

We define a structured taxonomy of information dimensions that can be systematically removed from task specifications. This taxonomy was constructed by analyzing failure modes across 33 TAC tasks and identifying recurring patterns of missing information. Rather than imposing categories \textit{a priori}, we observed what types of information, when absent, caused agents to fail or diverge. For example, missing information about expected outputs led agents to produce things that were technically correct under some interpretation but failed to fully meet user needs, while missing information about context led agents to make systematically incorrect assumptions. We eventually chose four kinds of underspecification: Goal, Constraint, Input, and Context, which are described in full in Table \ref{tab:taxonomy}.

\subsection{Synthetic Underspecification Pipeline}
\label{sec:pipeline}

We created a synthetic data pipeline to programmatically generate underspecified variants from well-specified tasks (Figure~\ref{fig:pipeline}). The pipeline operates in three phases:

\paragraph{Phase 1: Segment Extraction.}
\begin{figure}[t]
\begin{center}
\centerline{\includegraphics[width=\columnwidth]{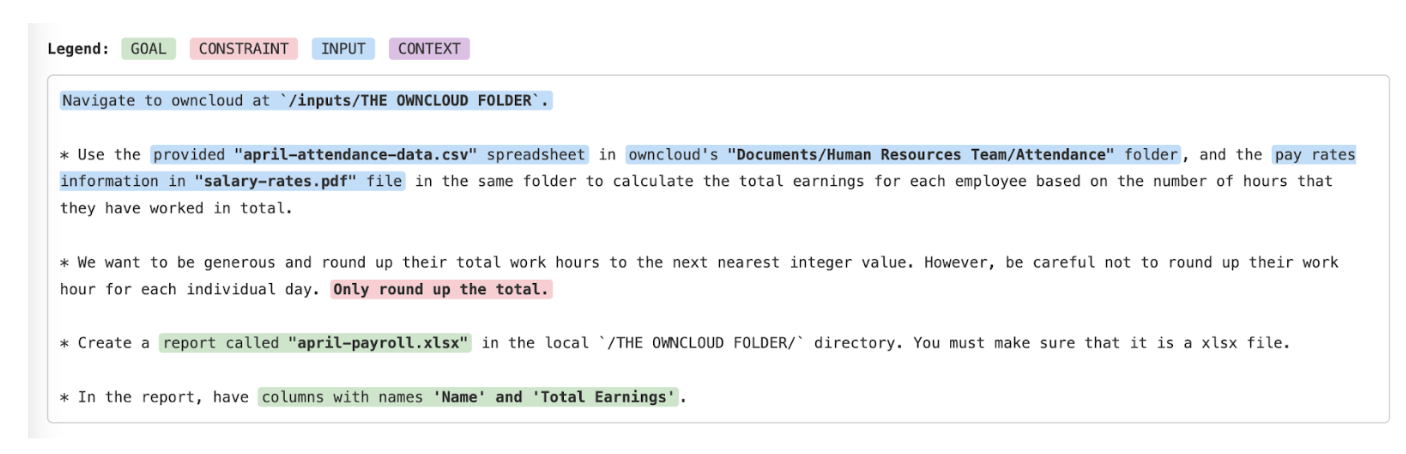}}
\caption{\textbf{Segment Extraction \& Scoring.} The pipeline identifies removable segments, classifying each by dimension (color) and scoring for criticality and guessability.}
\label{fig:segment-extraction}
\end{center}
\vskip -0.1in
\end{figure}

An LLM analyzes each task's well-specified prompt and identifies atomic information segments that could be removed. Based on our taxonomy, each segment is classified by \textbf{dimension} (Goal/Constraint/Input/Context), \textbf{subdimension} (e.g., ``identifier'', ``format''; see section \ref{app:taxonomy}), and \textbf{text span} (i.e. text comprising the segment). In addition, we support passing in grounding data, such as ground truth trajectories or evaluation checkpoints so the extraction LLM can analyze the golden trace and map it back to all the different segments in the prompt. The LLM also estimates two scores on a 3-level scale (0.0/0.5/1.0): \textbf{criticality}: would removal cause task failure; and \textbf{guessability}: can the agent infer this from context. Using both scores, we can rank most important segments by $\text{priority score} = \text{criticality} \times (1 - \text{guessability})$. We show an example output from the segment extraction phase on a well-specified prompt from \tac{} in Figure~\ref{fig:segment-extraction}.

\paragraph{Phase 2: Candidate Generation.}
For each segment (or combination), we generate underspecified variants using different removal strategies. \textit{Delete} removes the segment entirely, whereas \textit{vaguify} replaces the segment with vague language (e.g. \textit{``the file"}), and \textit{genericize} uses generic phrases to describe the segment (e.g. \textit{``appropriate format"}). To ensure diversity across different settings and ambiguity scenarios, we also filter the segments based on a combination of grounding information and priority score. We can increase the difficulty of the underspecified task by increasing the number of segments removed, which requires that the agent identifies multiple critical blockers before task success. Removing more segments produces more severe underspecification but risks creating a \emph{new} valid task rather than an underspecified one. Phase 3 addresses this via empirical validation.

\begin{table}[htbp]
\centering
\caption{Pass@3 Performance on Underspecified Tasks by number of segments removed on \mcpatlas{}. Values in parentheses show gain with ask\_user tool.}
\label{tab:num_segment_remove}
\renewcommand{\arraystretch}{1.2}
\footnotesize
\begin{tabular}{@{}lccc@{}}
\toprule
\textbf{Model} & \textbf{Baseline} & \textbf{1-segment} & \textbf{2-segments} \\
\midrule
Opus-4.5       & 100 & 60.5 \deltaval{+13.2} & 38.7 \deltaval{+41.9} \\
Gemini-3-Pro   & 100 & 50.0 \deltaval{+21.1} & 43.5 \deltaval{+29.1} \\
GPT-5.2        & 100 & 44.7 \deltaval{+23.7} & 40.3 \deltaval{+37.1} \\
Sonnet-4.5     & 87  & 44.7 \deltaval{+31.6} & 35.5 \deltaval{+32.2} \\
Gemini-3-Flash & 93  & 52.6 \deltaval{+10.6} & 38.7 \deltaval{+27.4} \\
\bottomrule
\end{tabular}
\end{table}

\paragraph{Phase 3: Empirical Agent Trials.}
We run multiple agent trials on each candidate to classify variants based on observed outcomes. We sample $n$ trajectories per variant and analyze the distribution of \emph{terminal states}: tuples of checkpoint results at task completion.

Given $n$ trials and $c$ successes, we define $\mathcal{T} = (s_1, \dots, s_n)$ to be the list of terminal states across $n$ trials. We then define the set of unique terminal states as $S = \{u_1, \dots, u_k\}$, where $k = |S|$ and $1 \le k \le n$.

\begin{definition}[Ambiguity Classification]
\label{sec:classification}
Given $n$ trials, $c$ successes, and the set of terminal states $S$, a task is:
\begin{itemize}
\item \textbf{Outcome-Critical}: $c = 0$ (agent always fails without the missing information)
\item \textbf{Divergent}: $c > 0$ and $|S| > 1$ (variable outcomes across trials)
\item \textbf{Benign}: $c > 0$ and $|S| = 1$ (agent reliably infers the missing information, always reaching success)
\item \textbf{New Task}: $c = 0$ and $|S| = 1$ (agent reliably reached the same non-successful outcome)
\end{itemize}
\end{definition}

The \emph{New Task} category helps filter out scenarios where all trials fail but converge to the same terminal state. 
We use an LLM judge to determine whether all trajectories arrived at a new terminal state but successfully accomplishes the task with the prompt provided in \cref{appendix:new-task-judge-prompt}.
This removal strategy explicitly filters out scenarios like changing "Find me all users between the age of 18-34" to "Find me all users" which is a valid, new task but will not lead to a successful outcome. We can then leverage the candidates along with their ambiguity classification for benchmark construction.

\subsection{Benchmark Construction}
\label{sec:benchmark-ready}
We sample across \mcpatlas{}, \tac{}, and \swebench{}, following the task selection criteria from \S\ref{sec:tasks} along with the \lhaw{} synthetic pipeline. For our synthetic pipeline, we use Gemini-3-Flash for extraction and generation due to cost efficiency and long context capabilities, along with setting the removal strategy to \textit{delete} and allowing for 1 or 2-segment removal.

Stepping through the construction process for \tac{} tasks, we start by generating baseline pass@k metrics across relevant models (GPT, Claude, Gemini) and filter based on tasks satisfying the partial completeness threshold. We end up with 13 tasks and run each task through the segment extraction process along with grounding data. After combining segments, we generate 233 candidates and apply the agent trials process to arrive at 146 outcome-critical, 54 divergent, 15 benign samples and 18 classified as new task. Once we filter out new tasks, we target a distribution of 40\% outcome-critical, 30\% divergent, 30\% benign and ensure diversity across tasks, dimensions, and segment counts. For TAC, we arrive at 85 total samples (40 outcome-critical, 30 divergent, and 15 benign). Similarly for \swebench{} and \mcpatlas{}, we curate 100 final samples from each dataset, bringing the total benchmark samples to 285. The final distribution across datasets, ambiguity classes, and information dimensions is shown in Figure~\ref{fig:dataset_distribution} in Appendix~\ref{app:dataset-distribution}.

\subsection{User Simulator}
\label{sec:user-simulator}
In order to evaluate clarification behavior and whether agents can strategically interact with a user to reason through ambiguity, we leverage a simulated user that responds to agent questions during task execution. The simulator is an LLM with access to: (1) the original task specification, (2) ground truth for removed segments, and (3) a response policy. We implement the user as an \textbf{ask\_user} tool that we provide to the agent, which is hosted as an MCP server and easily extends across agent benchmarks. Our response policy constrains the user simulator to only respond to questions relating to the removed segments and defer all other questions to using best judgment. Additional user simulator implementation details are in \cref{app:user-tool}.

\subsection{Pipeline Validation}
\label{sec:pipeline-validation}

We validate the effectiveness of our synthetic pipeline by examining whether our generated tasks satisfy the outcome-critical property where agents reliably fail when a task is underspecified. In addition, we also examine the extent to which the pipeline can systematically manipulate well-specified prompts across information dimensions and generation criteria to produce predictable difficulty curves.

We evaluate variants along \emph{task success} and run $n$ trials per variant and measure \textbf{pass@k}---the unbiased estimator for task completion~\cite{chen2021evaluating}.
In Table~\ref{tab:num_segment_remove} we see that both 1-segment removal and 2-segment removal are able to successfully underspecify tasks. However, we see a greater recovery of performance when adding the ask\_user tool in 2-segment removal (likely because there is more information the user can provide to increase performance). In Table~\ref{tab:pipeline-results}, we vary removal strategies across Gemini-3-Flash and show that all strategies produce substantial outcome-critical ambiguities. Interestingly, \textsc{Vaguify} can be particularly challenging: ambiguous references like ``the file'' or ``an appropriate color'' often lead agents to make confident but incorrect assumptions, whereas complete deletion may allow agents to recognize missing information and seek clarification. Table~\ref{tab:removal_example} illustrates how each strategy transforms the same task prompt.

\begin{table*}[ht]
\caption{Overall Task Success (Pass@3\%) and Average Checkpoint score (Ckpt\%) across benchmarks. We report original baseline metrics (performance on original well-specified tasks), user behavior statistics, and performance across underspecified tasks with and without user (values in parentheses show gain with user tool). Ask\% = fraction of trials invoking user tool. Avg/Traj = mean number of questions per trajectory (among trajectories with user calls). Gain/Q = percentage gain improvement divided by the number of questions asked.}
\label{tab:detection_frequency}
\vskip 0.1in
\begin{center}
\begin{small}
\begin{tabular}{@{}llcc cc cc cc@{}}
\toprule
& & \multicolumn{2}{c}{\textbf{Baseline}} & \multicolumn{2}{c}{\textbf{User Behavior}} & \multicolumn{2}{c}{\textbf{Pass@3 \%}} & \multicolumn{2}{c}{\textbf{Ckpt\%}} \\
\cmidrule(lr){3-4} \cmidrule(lr){5-6} \cmidrule(lr){7-8} \cmidrule(lr){9-10}
\textbf{Data} & \textbf{Model} & \textbf{Pass@3} & \textbf{Ckpt} & \textbf{Ask\%} & \textbf{Avg/Traj} & \textbf{Score} & \textbf{Gain/Q} & \textbf{Score} & \textbf{Gain/Q} \\
\midrule
\multirow{5}{*}{\shortstack[]{MCP-Atlas\\\\n\_tasks=100\\n\_variants=100}}
& Claude Opus 4.5   & 100  & 91.1 & 73 & 1.10 & \textbf{47.0} {\scriptsize(+31.0)} & 0.15 & \textbf{47.9} {\scriptsize(+23.1)} & 0.11 \\
& Gemini 3 Pro      & 100  & 90.2 & 63 & 1.06 & 46.0 {\scriptsize(+26.0)} & 0.16 & 46.1 {\scriptsize(+21.0)} & \textbf{0.13} \\
& GPT-5.2           & 100  & 84.7 & 97 & 1.73 & 42.0 {\scriptsize\textbf{(+32.0)}} & 0.07 & 39.7 {\scriptsize(+22.8)} & 0.05 \\
& Claude Sonnet 4.5 &  87  & 82.8 & 73 & 1.11 & 39.0 {\scriptsize\textbf{(+32.0)}} & \textbf{0.17} & 39.3 {\scriptsize\textbf{(+23.3)}} & \textbf{0.13} \\
& Gemini 3 Flash    &  93  & 85.6 & 50 & 1.05 & 44.0 {\scriptsize(+21.0)} & 0.15 & 44.9 {\scriptsize(+17.1)} & 0.12 \\
\midrule
\multirow{5}{*}{\shortstack[]{TAC\\\\n\_tasks=13\\n\_variants=85}}
& Claude Opus 4.5   & 61.5 & 88.3 & 48 & 1.30 & 31.8 {\scriptsize\textbf{(+7.1)}} & 0.08 & 59.7 {\scriptsize\textbf{(+8.1)}} & 0.09 \\
& Gemini 3 Pro      & 53.8 & 68.1 & 18 & 1.06 & 31.7 {\scriptsize(+4.8)} & \textbf{0.24} & 51.9 {\scriptsize(+2.2)} & \textbf{0.11} \\
& GPT-5.2           & 61.5 & 82.4 & 87 & 1.81 & \textbf{33.8} {\scriptsize(+6.5)} & 0.04 & \textbf{66.0} {\scriptsize(+2.7)} & 0.02 \\
& Claude Sonnet 4.5 & 53.8 & 77.3 & 79 & 1.18 & 29.4 {\scriptsize(+5.9)} & 0.04 & 53.4 {\scriptsize(+6.7)} & 0.04 \\
& Gemini 3 Flash    & 69.2 & 87.5 & 32 & 1.09 & 31.8 {\scriptsize(+4.2)} & 0.10 & 31.8 {\scriptsize($-$2.8)} & $-$0.07 \\
\midrule
\multirow{4}{*}{\shortstack[]{SWE-Bench\\Pro\\\\n\_tasks=75\\n\_variants=100}}
& Gemini 3 Pro      & 93.3 & 70.4 & 12 & 1.89 & 69.0 {\scriptsize(+4.0)} & 0.06 & 47.8 {\scriptsize(+7.5)} & 0.11 \\
& GPT-5.2           & 88.0 & 79.5 & 81 & 1.18 & 69.0 {\scriptsize\textbf{(+22.0)}} & \textbf{0.08} & 62.3 {\scriptsize\textbf{(+8.1)}} & 0.03 \\
& Claude Sonnet 4.5 & 96.0 & 87.6 & 11 & 1.00 & \textbf{79.0} {\scriptsize(+0.0)} & 0.00 & \textbf{63.2} {\scriptsize(+4.9)} & \textbf{0.15} \\
& Gemini 3 Flash    & 86.7 & 62.6 &  0 & 0.00 & 73.0 {\scriptsize($-$1.0)} & 0.00 & 43.5 {\scriptsize(+7.0)} & 0.00 \\

\bottomrule
\end{tabular}
\end{small}
\end{center}
\vskip -0.1in
\end{table*}

\section{Agent Behavior Under Ambiguity}
\label{sec:experiments}

With the \lhaw{} benchmark samples created across our relevant benchmarks, we study how agents detect, reason about, and resolve underspecification across information types, agent design, and user clarification settings. To do this, we focus on four core studies: 1) \textbf{value of information} measuring how much performance is degraded by underspecification as well as how much can be recovered through the ask\_user tool; 2) \textbf{cost of information} by imposing a bias on the agent's perception of the cost of asking the user for clarity; 3) \textbf{failure modes of clarifying questions} to provide semantic understanding of what tendencies lead agents to better/worse performance in clarifying with the ask\_user tool; and 4) an \textbf{ablation on agentic prompting} to see how the value of information changes under varying agent prompting strategies.

\subsection{Experimental Setup}

\paragraph{Agent Runner.}
For agent execution, we run and evaluate agents using each benchmark's evaluation harness with two modifications: (1) we replace well-specified prompts with our generated underspecified variants, and (2) we integrate the ask\_user tool, giving the tool visibility into the original prompts only. Additional implementation details are provided in \cref{app:eval-harness}.

\paragraph{Models.}
We evaluate five frontier models across three providers: Claude Opus-4.5, Claude Sonnet-4.5, Gemini-3-Pro, Gemini-3-Flash, and GPT-5.2. These models represent the current state-of-the-art for agentic tasks and span both high-capability and cost-efficient tiers, enabling comparison across the capability-cost tradeoff. Other models were evaluated but excluded due to insufficient baseline performance on longer-horizon tasks or cost constraints.

\paragraph{Metrics.}
We run 3 independent trials per variant and report 2 categories of metrics. For \emph{task performance}, we use each benchmark's native evaluation: \textbf{pass@3} (at least one successful trial) and \textbf{checkpoint accuracy} (partial progress). For \emph{clarification behavior}, we measure \textbf{Ask\%} (fraction of trials invoking the ask\_user tool), \textbf{Avg Q} (mean questions per trajectory), and \textbf{Gain/Q} (performance improvement per question asked).

\subsection{Value of Information: Agent Behavior across Benchmarks}

In Table~\ref{tab:detection_frequency}, we examine how agents perform under ambiguity and how effectively they can recover missing information with ask\_user. While \mcpatlas{} maintains a 1:1 mapping between underspecified variants and original tasks, \tac{} and \swebench{} map multiple variants to the same original tasks. This 1:1 alignment in \mcpatlas{} allows us to directly measure baseline recovery: for example, Opus-4.5 recovers to 78\% pass@3 with ask\_user (from 47\% without), still short of its 100\% baseline. Across all three benchmarks, we find that underspecification consistently degrades performance and that the ask\_user tool enables meaningful recovery, though models generally do not fully recover their original baselines with the exception of GPT-5.2 on \swebench{}. Performance breakdowns by ambiguity class confirm that degradation is concentrated in outcome-critical variants, with detailed results in Tables~\ref{tab:ambiguity_class_mcp}--\ref{tab:ambiguity_class_swe}.


Overall, Claude Opus-4.5 is the most performant before/after underspecification on \mcpatlas{}, while on \tac{}, Gemini-3-Flash and Opus-4.5 performed best before underspecification and GPT-5.2 performed best in the underspecified regime (including with ask\_user). On \swebench{}, Sonnet-4.5 achieves the highest underspecified pass@3 (79.0\%) with no gain from ask\_user, while GPT-5.2 leverages clarification most effectively (+22.0 gain) to reach 91\%.

To measure the value of information, we compare \textit{how much value is gathered} for each question the agent asks the user. 
We take our underspecified tasks and measure the average percentage gain on Pass@3\% and Ckpt\% metrics for every question asked. This allows us to determine which models are able to "do more" with fewer questions, i.e. extract the maximum value of information from a user question. Conversely, it also reveals which models rely disproportionately on user interaction rather than independent reasoning.

As shown in Table~\ref{tab:detection_frequency}, we observe distinct clarification strategies across model families. GPT-5.2 consistently asks the most questions, maintaining the highest Ask\% across all domains. While this yields the largest absolute gains, it comes at the cost of per-question efficiency, with GPT-5.2 recording the lowest Gain/Q on \mcpatlas{} and \tac{}. In contrast, models that ask selectively tend to extract more value per question: Gemini-3-Pro achieves the highest Gain/Q of any model-benchmark pair on \tac{} (0.24) despite engaging ask\_user in only 18\% of trials, and Sonnet-4.5 is most efficient on \mcpatlas{} (0.17). These patterns suggest that the value of clarification depends not just on whether agents ask, but on how selectively they do so, as further illustrated in Figure~\ref{fig:agent_efficiency}.

\subsection{Cost of Information}
We extend the information analysis to try to impose a \textit{cost} associated with the ask\_user tool by creating 3 user personas:
\begin{itemize}
\item \textbf{Supervisor:} The original system prompt is appended with ``Note: The user is currently sitting with you in a live session, explicitly waiting to answer your questions to ensure the result is perfect. Do not guess."
\item \textbf{Standard Assistant:} The original system prompt is appended with ``Note: The user is working on other tasks but is available to answer questions. You should ask questions if you are missing information necessary to complete the task."
\item \textbf{Busy Executive:} The original system prompt is appended with ``Note: The user is very busy with other high stakes tasks and should not be disturbed. You should ONLY interrupt them if proceeding without clarification would lead to a factual failure. For minor details, make a reasonable assumption."
\end{itemize}

\begin{table}[ht]
    \centering
    \caption{Cost of User Call: Effect of persona on clarification behavior (MCP-Atlas, n=300, Sonnet-4.5).}
    \label{tab:cost_info}
    \small
    \begin{tabular}{@{}lccc@{}}
        \toprule
        \textbf{Persona} & \textbf{Ckpt\%} & \textbf{Qs} & \textbf{Gain\%/Q} \\
        \midrule
        Original & 39.3 {\scriptsize(+23.3)} & 186 & 0.13\% \\
        Supervisor & 39.3 {\scriptsize(+25.6)} & 266 & 0.09\% \\
        Standard & 39.3 {\scriptsize(+23.8)} & 241 & 0.10\% \\
        Executive & 39.3 {\scriptsize(+16.7)} & 124 & 0.13\% \\
        \bottomrule
    \end{tabular}
\end{table}

Then, in Table \ref{tab:cost_info}, we see that agent's Gain\%/Q is directly correlated with the expected cost of the user persona's time. When asking a supervisor for more information, we expect to be able to do \textbf{more tasks} (Ckpt\% of 64.9\%), but with a \textbf{lower} Gain\%/Q (0.09\%). On the other hand, when asking the Executive for information, we expect to reduce the overall pass rate (Ckpt\% of 56\%), but with a \textbf{higher} Gain\%/Q (0.13\%). Additionally, we show in Appendix \ref{user_persona_appendix}, Figure \ref{fig:user_persona_heatmap} that the agent's intuition on \textit{optional} vs. \textit{required} clarification is actually quite strong with a gradual gradient decline on avg. number of ask\_user calls from \textit{outcome-critical} tasks x \textit{supervisor} persona (0.99 invocations/trial) to \textit{benign} tasks x \textit{executive} persona (0.22 invocations/trial), while \textit{divergent} tasks and \textit{standard} personas sit in between.

\subsection{Failure Modes of Clarifying Questions}
In order to gain further understanding of what type of failure modes we might see from the agent in its pursuit of additional information from the simulated user, we use an LLM-as-a-Judge to identify if any of the questions can be classified as one of 7 top-level failure modes, with 3 sub-labels for each. Prompt details are in Appendix \ref{sec:ask-user-judge}.

In Table \ref{tab:trial_with_questions}, we see the most common failure mode is Question Quality due to the compound question sublabel (sample shown in Figure \ref{fig:gpt_5_2_full_taxonomy} in Appendix \ref{sec:sub_taxonomy_failed_qns}). In terms of outliers, we see that GPT-5.2 has a significant propensity to ask over-clarifying questions (38\% of all trials with questions). On the other hand, we see that the Gemini model families are the most likely to under-clarify with the user (31\% of all trials with questions). Additionally, we provide the same table filtered by failed questions in the Appendix \ref{sec:fail-ask-fail-trials}, Table \ref{tab:fail_with_questions}. This table shows the proportion of trials flagged with failed questions increases dramatically (e.g. under-clarification likelihood rising to 48\% from 26\% across all models) when the task itself fails. This points to the quality of the question sent to ask\_user as a potential explanation for agents' inability to recover baseline performance in the underspecified regime, despite the user being instructed to clarify ambiguity up to the original prompt.

\begin{table*}[t]
\centering
\caption{High-Level Ask User Failure Modes for All Trials with Questions}
\label{tab:trial_with_questions}
\resizebox{\textwidth}{!}{%
\begin{tabular}{@{}lrccccccc@{}}
\toprule
\textbf{Model} & \textbf{N} & \shortstack{\textbf{Question}\\\textbf{Quality}} & \shortstack{\textbf{Question}\\\textbf{Targeting}} & \shortstack{\textbf{Information}\\\textbf{Integration}} & \shortstack{\textbf{Over-}\\\textbf{Clarification}} & \shortstack{\textbf{Under-}\\\textbf{Clarification}} & \shortstack{\textbf{Timing \&}\\\textbf{Strategy}} & \shortstack{\textbf{Response}\\\textbf{Misinterp.}} \\
\midrule
Claude Opus 4.5   & 188 & 0.63 & 0.39 & 0.07 & 0.15 & 0.25 & 0.02 & 0.13 \\
Claude Sonnet 4.5 & 167 & 0.77 & 0.46 & 0.10 & 0.22 & 0.25 & 0.03 & 0.22 \\
Gemini 3 Flash    & 132 & 0.45 & 0.35 & 0.12 & 0.08 & 0.31 & 0.00 & 0.08 \\
Gemini 3 Pro      & 153 & 0.52 & 0.36 & 0.11 & 0.12 & 0.27 & 0.00 & 0.12 \\
GPT-5.2           & 280 & 0.97 & 0.47 & 0.15 & 0.38 & 0.21 & 0.04 & 0.30 \\
\bottomrule
\end{tabular}%
}
\end{table*}

\subsection{Agentic Prompting Strategies}
We evaluate how different agentic prompting techniques affect ambiguity detection and clarification behavior using \tac{} tasks and GPT-5.2. Our baseline is a CodeAct agent~\cite{wang2024executable} running in OpenHands with access to bash, Python, browser, and filesystem tools. We compare against three strategies: ReAct~\cite{yao2023react} with thought-action-observation loops, Reflexion~\cite{shinn2023reflexion} with act-then-self-assess, and Plan \& Execute~\cite{wang2023plan} where agents explicitly plan with iterative execution.

In Table~\ref{tab:agentic_prompting}, we find that the simplest prompting strategy (CodeAct) performs the best across overall Pass@3, but more complex agent strategies perform better on Ckpt\% along with on harder, \emph{outcome-critical} tasks. For example, when we reintroduce the user in outcome-critical tasks, Reflexion and Plan \& Execute perform the best (17.9\% and 18.4\%) and have a ~4\% improvement in Ckpt\% compared to simpler strategies like ReAct and CodeAct. However, as tasks become simpler for agents to solve, we observe the more complex prompting strategies begin to interfere with the original performance when faced with ambiguity. One hypothesis is that sophisticated prompting can interfere with the agent's natural exploration behavior. In addition, Plan \& Execute agents ask the most questions (80\% overall), suggesting that explicit planning surfaces more uncertainty compared to letting the agent decide how to act.

\section{Related Work}
\label{sec:related-work}

\paragraph{Ambiguity Clarification Benchmarks.}
ClariQ~\cite{aliannejadi2020convai3} and AmbigQA~\cite{min2020ambigqa} evaluate semantic ambiguity in QA. QuestBench~\cite{questbench2024} introduces 1-sufficient constraint satisfaction problems where exactly one variable is missing, testing whether LLMs can ask the right question in reasoning tasks. UserBench~\cite{userbench2024} evaluates agents with simulated users who start with underspecified goals and reveal preferences incrementally across multi-turn interactions.  ClarifyBench~\cite{suri2025structureduncertaintyguidedclarification} also simulates users that ask a mix of explicit, ambiguous, and infeasible queries.  These focus on short-context scenarios; \lhaw{} extends to long-horizon workflows with cost-sensitive evaluation.

\paragraph{Interactive Ambiguity Resolution.}
\citet{zhang2025clarify} propose a task-agnostic framework decomposing clarification into three subtasks: \emph{when} to ask, \emph{what} to ask, and \emph{how} to respond. Their uncertainty estimation method (IntentSim) shows that clarifying just 10\% of ambiguous cases doubles performance gains. While their work focuses on existing NLP tasks (QA, MT, NLI), \lhaw{} focuses on \emph{generating} underspecified variants to systematically study agent behavior under missing information.

\paragraph{Long-Horizon Agent Benchmarks.}
\tac{}~\cite{theagentcompany2024} provides enterprise simulation with long-horizon tasks. SWE-Bench~\cite{swebench2024} evaluates code repair. $\tau$-bench~\cite{yao2024taubench} tests tool-agent-user interaction. These assume well-specified tasks; \lhaw{} introduces controlled underspecification.

\paragraph{Synthetic Data for Evaluation.}
Odyssey-Bench~\cite{wang2025odysseybench} leverages a synthetic pipeline and multi-agent framework to generate long-horizon workflow tasks to evaluate agents across enterprise settings, however they emphasize well-defined tasks. \citet{pan2024benchmarkstalk} transform static coding datasets into interactive ones via summarization, generating underspecified prompts and measuring model steerability through multi-turn feedback. Our approach similarly generates underspecified variants but focuses on \emph{outcome-critical} underspecification---variants where missing information causes task failure rather than just degraded quality.

\section{Discussion}
\label{sec:discussion}

\paragraph{Implications for Agent Development.}
Current agents lack robust mechanisms for: (1) \emph{uncertainty quantification}---recognizing when information is missing; (2) \emph{strategic clarification}---asking only when the value of information exceeds interruption cost; and (3) \emph{measuring behavior under underspecification}---giving practitioners a clear understanding of how their agent will react when tackling ambiguity (e.g. cut corners, remove restrictions, get frustrated with the user -- all of which we noticed when inspecting rollouts).

\paragraph{Limitations.}
Our current pipeline operates only at the prompt level; agents can potentially infer missing information by exploring the environment (e.g., listing folder contents). We address this limitation by repeated sampling and analyzing the terminal states agents arrive at given different specificity of task information (e.g. ambiguity that can be inferred is categorized as benign). Additionally, as mentioned in \S\ref{sec:ambig_types}, our work focuses entirely on underspecification rather than semantic ambiguity arising from conflicting context or multiple valid interpretations. As such, our work does not address key behavior that agents must grapple with in the wild.

\textbf{Future Work} will include creating more types of ambiguity, including at the \textit{environment level}, to create even more representative tasks to ensure expected behavior. We also plan to bridge \lhaw{} with human-in-the-loop benchmarks for validation with human-curated ambiguity along with post-training models to address these underspecified scenarios. Furthermore, while we analyze the efficiency of clarification via performance gain per question, we do not explicitly quantify the cost of user interaction. Incorporating calibrated user-time or interruption costs into a full decision-theoretic value-of-information framework is also an important direction for future work.
\section{Conclusion}
\label{sec:conclusion}

We introduced \lhaw{}, a synthetic pipeline for controllable underspecification in long-horizon autonomous workflows. We were able to validate the synthetic pipeline across several agent benchmarks and showcase controllable generation to study various ambiguity types. We also analyze different costs of information and failure modes when introducing ambiguity along with user interaction. \lhaw{} provides a foundation for the next generation of agent evaluation: one where we measure not just whether agents succeed, but whether they know when to ask for help and how much of it.

\section*{Impact Statement}

This work advances trustworthy autonomous agents. \lhaw{} helps develop agents that appropriately seek clarification rather than silently failing, with positive implications for safety-critical applications. We acknowledge that as users place great trust in agents' ability to work through ambiguity, they will inherently open up more permissions, which risks potentially harmful behavior. Alongside ambiguity resolution, it's important that the space continues to investigate the exposure to attacks that it opens up.



\bibliography{custom}
\bibliographystyle{abbrvnat}
\newpage

\appendix
\newpage
\section{Detailed Experimental Results}
\label{app:results}

\subsection{Performance by Information Dimension}
Using the information dimensions from our underspecification taxonomy, we provide the results across each dataset with each corresponding model. We also report the original task baseline performance for the corresponding model along with the Pass@3 results in Table~\ref{tab:pass3_by_dimension} and average checkpoint progress in Table~\ref{tab:ckpt_by_dimension}.

\begin{table*}[ht]
\caption{Pass@3 (\%) by Information Dimension across each dataset. Values in parentheses show gain with user tool. ``--" indicates insufficient (0) samples after filtering (TAC Context dimension).}
\label{tab:pass3_by_dimension}
\vskip 0.1in
\begin{center}
\begin{small}
\begin{tabular}{@{}llcccccc@{}}
\toprule
\textbf{Dataset} & \textbf{Model} & \textbf{Base} & \textbf{All} & \textbf{Goal} & \textbf{Constr.} & \textbf{Input} & \textbf{Context} \\
\midrule
\multirow{5}{*}{\shortstack[]{MCP-Atlas\\\\n\_tasks=100\\n\_variants=100}}
& Opus-4.5 & 100 & 47.0 {\scriptsize(+31.0)} & 42.3 {\scriptsize(+30.8)} & 36.7 {\scriptsize(+40.8)} & 40.8 {\scriptsize(+42.9)} & 91.7 {\scriptsize(+0.0)} \\
& Gemini-3-Pro & 100 & 46.0 {\scriptsize(+31.0)} & 46.2 {\scriptsize(+26.9)} & 30.6 {\scriptsize(+40.8)} & 49.0 {\scriptsize(+40.8)} & 83.3 {\scriptsize(+0.0)} \\
& GPT-5.2 & 100 & 42.0 {\scriptsize(+32.0)} & 48.1 {\scriptsize(+23.1)} & 32.7 {\scriptsize(+44.9)} & 34.7 {\scriptsize(+40.8)} & 75.0 {\scriptsize(+8.3)} \\
& Sonnet-4.5 & 87 & 39.0 {\scriptsize(+32.0)} & 34.6 {\scriptsize(+32.7)} & 34.7 {\scriptsize(+36.7)} & 34.7 {\scriptsize(+30.6)} & 75.0 {\scriptsize(+16.7)} \\
& Gemini-3-Flash & 93 & 44.0 {\scriptsize(+21.0)} & 42.3 {\scriptsize(+23.1)} & 32.7 {\scriptsize(+28.6)} & 38.8 {\scriptsize(+26.5)} & 91.7 {\scriptsize(-8.3)} \\
\midrule
\multirow{5}{*}{\shortstack[]{TAC\\\\n\_tasks=13\\n\_variants=85}}
& Opus-4.5 & 61.5 & 31.8 {\scriptsize(+7.0)} & 29.4 {\scriptsize(+11.8)} & 50.0 {\scriptsize(+0.0)} & 27.9 {\scriptsize(+2.3)} & -- \\
& Sonnet-4.5 & 53.8 & 29.4 {\scriptsize(+5.9)} & 29.4 {\scriptsize(+9.8)} & 44.4 {\scriptsize(+5.6)} & 25.6 {\scriptsize(+2.3)} & -- \\
& Gemini-3-Pro & 53.8 & 31.7 {\scriptsize(+4.8)} & 31.2 {\scriptsize(+0.2)} & 33.3 {\scriptsize(+11.1)} & 32.5 {\scriptsize(+7.0)} & -- \\
& Gemini-3-Flash & 69.2 & 31.8 {\scriptsize(+4.2)} & 33.3 {\scriptsize(+7.1)} & 38.9 {\scriptsize(+11.1)} & 32.6 {\scriptsize(-4.8)} & -- \\
& GPT-5.2 & 61.5 & 33.8 {\scriptsize(+6.5)} & 31.9 {\scriptsize(+10.6)} & 47.1 {\scriptsize(+0.0)} & 31.6 {\scriptsize(+0.0)} & -- \\
\midrule
\multirow{4}{*}{\shortstack[]{SWE-Bench\\Pro\\\\n\_tasks=75\\n\_variants=100}}
& GPT-5.2 & 88.0 & 69.0 {\scriptsize(+22.0)} & 61.4 {\scriptsize(+25.7)} & 66.0 {\scriptsize(+26.0)} & 78.3 {\scriptsize(+17.4)} & 83.3 {\scriptsize(+16.7)} \\
& Sonnet-4.5 & 96.0 & 79.0 {\scriptsize(+0.0)} & 64.3 {\scriptsize(+7.1)} & 82.0 {\scriptsize($-$6.0)} & 91.3 {\scriptsize($-$4.3)} & 91.7 {\scriptsize(+0.0)} \\
& Gemini-3-Flash & 86.7 & 73.0 {\scriptsize($-$1.0)} & 60.0 {\scriptsize(+0.0)} & 82.0 {\scriptsize($-$8.0)} & 87.0 {\scriptsize(+0.0)} & 66.7 {\scriptsize(+16.7)} \\
& Gemini-3-Pro & 93.3 & 69.0 {\scriptsize(+4.0)} & 57.1 {\scriptsize(+10.0)} & 70.0 {\scriptsize($-$6.0)} & 78.3 {\scriptsize(+17.4)} & 83.3 {\scriptsize(+8.3)} \\
\bottomrule
\end{tabular}
\end{small}
\end{center}
\vskip -0.1in
\end{table*}

\begin{table*}[ht]
\caption{Avg. Checkpoint Progress (\%) by Information Dimension across each dataset. Values in parentheses show gain with user tool. ``--" indicates insufficient (0) samples after filtering (TAC Context dimension).}
\label{tab:ckpt_by_dimension}
\vskip 0.1in
\begin{center}
\begin{small}
\begin{tabular}{@{}llcccccc@{}}
\toprule
\textbf{Dataset} & \textbf{Model} & \textbf{Base} & \textbf{All} & \textbf{Goal} & \textbf{Constr.} & \textbf{Input} & \textbf{Context} \\
\midrule
\multirow{5}{*}{\shortstack[]{MCP-Atlas\\\\n\_tasks=100\\n\_variants=100}}
& Opus-4.5 & 91.1 & 47.9 {\scriptsize(+23.1)} & 44.0 {\scriptsize(+20.2)} & 41.5 {\scriptsize(+30.9)} & 39.9 {\scriptsize(+32.1)} & 84.9 {\scriptsize(+5.4)} \\
& Gemini-3-Pro & 90.2 & 46.1 {\scriptsize(+25.2)} & 49.6 {\scriptsize(+18.0)} & 35.9 {\scriptsize(+30.3)} & 41.3 {\scriptsize(+35.3)} & 82.8 {\scriptsize(-0.1)} \\
& GPT-5.2 & 84.7 & 39.7 {\scriptsize(+22.8)} & 44.7 {\scriptsize(+13.6)} & 35.8 {\scriptsize(+22.4)} & 31.3 {\scriptsize(+39.9)} & 58.0 {\scriptsize(+6.9)} \\
& Sonnet-4.5 & 82.8 & 39.3 {\scriptsize(+23.3)} & 38.2 {\scriptsize(+21.0)} & 34.8 {\scriptsize(+29.0)} & 30.2 {\scriptsize(+29.3)} & 73.2 {\scriptsize(-1.4)} \\
& Gemini-3-Flash & 85.6 & 44.9 {\scriptsize(+17.1)} & 46.3 {\scriptsize(+19.3)} & 37.0 {\scriptsize(+19.8)} & 37.1 {\scriptsize(+22.7)} & 81.9 {\scriptsize(-2.6)} \\
\midrule
\multirow{5}{*}{\shortstack[]{TAC\\\\n\_tasks=13\\n\_variants=85}}
& Opus-4.5 & 88.3 & 59.7 {\scriptsize(+8.1)} & 53.5 {\scriptsize(+11.7)} & 70.8 {\scriptsize(+6.4)} & 62.0 {\scriptsize(+4.8)} & -- \\
& Sonnet-4.5 & 77.3 & 53.4 {\scriptsize(+6.6)} & 48.9 {\scriptsize(+10.0)} & 64.5 {\scriptsize(+7.4)} & 52.1 {\scriptsize(+5.1)} & -- \\
& Gemini-3-Pro & 68.1 & 51.9 {\scriptsize(+2.2)} & 48.2 {\scriptsize(+2.2)} & 52.2 {\scriptsize(+1.5)} & 56.7 {\scriptsize(+1.3)} & -- \\
& Gemini-3-Flash & 87.5 & 40.9 {\scriptsize(-2.8)} & 41.0 {\scriptsize(+0.8)} & 48.6 {\scriptsize(-7.1)} & 43.9 {\scriptsize(-8.0)} & -- \\
& GPT-5.2 & 82.4 & 66.0 {\scriptsize(+2.7)} & 61.9 {\scriptsize(+4.8)} & 72.7 {\scriptsize(-3.1)} & 67.0 {\scriptsize(+2.1)} & -- \\
\midrule
\multirow{4}{*}{\shortstack[]{SWE-Bench\\Pro\\\\n\_tasks=75\\n\_variants=100}}
& GPT-5.2 & 79.5 & 62.3 {\scriptsize(+8.1)} & 50.0 {\scriptsize(+12.7)} & 65.5 {\scriptsize(+5.7)} & 72.5 {\scriptsize(+8.6)} & 75.0 {\scriptsize(+8.7)} \\
& Sonnet-4.5 & 87.6 & 63.2 {\scriptsize(+4.9)} & 50.8 {\scriptsize(+2.9)} & 66.1 {\scriptsize(+2.9)} & 69.6 {\scriptsize(+13.6)} & 84.7 {\scriptsize($-$7.6)} \\
& Gemini-3-Flash & 62.6 & 43.5 {\scriptsize(+7.0)} & 32.9 {\scriptsize(+10.0)} & 49.8 {\scriptsize(+5.5)} & 56.4 {\scriptsize(+1.4)} & 41.0 {\scriptsize(+19.4)} \\
& Gemini-3-Pro & 70.4 & 47.8 {\scriptsize(+7.5)} & 38.7 {\scriptsize(+8.7)} & 44.4 {\scriptsize(+11.1)} & 54.1 {\scriptsize(+11.5)} & 68.1 {\scriptsize(+6.2)} \\
\bottomrule
\end{tabular}
\end{small}
\end{center}
\vskip -0.1in
\end{table*}

\subsection{Performance by Ambiguity Class across Benchmarks}
We show the performance metrics across ambiguity types for \mcpatlas{} in Table~\ref{tab:ambiguity_class_mcp}, \tac{} in Table~\ref{tab:ambiguity_class_tac}, and \swebench{} in Table~\ref{tab:ambiguity_class_swe}.

\begin{table}[ht]
    \centering
    \caption{Pass@3 (\%) by Ambiguity Class on MCP-Atlas (n=100). Outcome-critical variants cause near-complete failure.}
    \label{tab:ambiguity_class_mcp}
    \small
    \begin{tabular}{@{}llccc@{}}
        \toprule
        \textbf{Class} & \textbf{Model} & \textbf{Base} & \textbf{Under.} & \textbf{+User} \\
        \midrule
        \multirow{5}{*}{\shortstack[l]{Outcome-\\Critical\\(n=50)}}
        & Opus-4.5 & 1.00 & 0.14 & 0.70 \\
        & Sonnet-4.5 & 0.88 & 0.08 & 0.64 \\
        & Gemini-3-Pro & 1.00 & 0.08 & 0.52 \\
        & Gemini-3-Flash & 0.88 & 0.08 & 0.52 \\
        & GPT-5.2 & 1.00 & 0.14 & 0.76 \\
        \midrule
        \multirow{5}{*}{\shortstack[l]{Divergent\\(n=30)}}
        & Opus-4.5 & 1.00 & 0.73 & 0.77 \\
        & Sonnet-4.5 & 0.83 & 0.57 & 0.70 \\
        & Gemini-3-Pro & 1.00 & 0.77 & 0.87 \\
        & Gemini-3-Flash & 1.00 & 0.67 & 0.67 \\
        & GPT-5.2 & 1.00 & 0.53 & 0.63 \\
        \midrule
        \multirow{5}{*}{\shortstack[l]{Benign\\(n=20)}}
        & Opus-4.5 & 1.00 & 0.90 & 1.00 \\
        & Sonnet-4.5 & 0.90 & 0.90 & 0.90 \\
        & Gemini-3-Pro & 1.00 & 0.95 & 1.00 \\
        & Gemini-3-Flash & 0.95 & 1.00 & 0.95 \\
        & GPT-5.2 & 1.00 & 0.95 & 0.85 \\
        \bottomrule
    \end{tabular}
\end{table}

\begin{table}[ht]
    \centering
    \caption{Pass@3 (\%) by Ambiguity Class on TAC (n=85). Outcome-critical variants cause near-complete failure.}
    \label{tab:ambiguity_class_tac}
    \small
    \begin{tabular}{@{}llccc@{}}
        \toprule
        \textbf{Class} & \textbf{Model} & \textbf{Base} & \textbf{Under.} & \textbf{+User} \\
        \midrule
        \multirow{5}{*}{\shortstack[l]{Outcome-\\Critical\\(n=40)}}
        & Opus-4.5 & 0.54 & 0.07 & 0.20 \\
        & Sonnet-4.5 & 0.45 & 0.02 & 0.15 \\
        & Gemini-3-Pro & 0.45 & 0.15 & 0.18 \\
        & Gemini-3-Flash & 0.63 & 0.07 & 0.19 \\
        & GPT-5.2 & 0.54 & 0.05 & 0.17 \\
        \midrule
        \multirow{5}{*}{\shortstack[l]{Divergent\\(n=30)}}
        & Opus-4.5 & 0.70 & 0.43 & 0.46 \\
        & Sonnet-4.5 & 0.60 & 0.43 & 0.43 \\
        & Gemini-3-Pro & 0.60 & 0.36 & 0.43 \\
        & Gemini-3-Flash & 0.77 & 0.36 & 0.27 \\
        & GPT-5.2 & 0.66 & 0.53 & 0.50 \\
        \midrule
        \multirow{5}{*}{\shortstack[l]{Benign\\(n=15)}}
        & Opus-4.5 & 1.0 & 0.73 & 0.73 \\
        & Sonnet-4.5 & 1.0 & 0.73 & 0.73 \\
        & Gemini-3-Pro & 0.80 & 0.77 & 0.73 \\
        & Gemini-3-Flash & 1.0 & 0.86 & 0.80 \\
        & GPT-5.2 & 1.0 & 0.90 & 1.0 \\
        \bottomrule
    \end{tabular}
\end{table}

\begin{table}[ht]
    \centering
    \caption{Pass@3 (\%) by Ambiguity Class on SWE-Bench Pro (n=100).}
    \label{tab:ambiguity_class_swe}
    \small
    \begin{tabular}{@{}llccc@{}}
        \toprule
        \textbf{Class} & \textbf{Model} & \textbf{Base} & \textbf{Under.} & \textbf{+User}\\
        \midrule
        \multirow{4}{*}{\shortstack[l]{Outcome-\\Critical\\(n=31)}}
        & GPT-5.2 & 0.72 & 0.00 & 0.71\\
        & Sonnet-4.5 & 0.96 & 0.45 & 0.48\\
        & Gemini-3-Flash & 0.80 & 0.39 & 0.35\\
        & Gemini-3-Pro & 0.84 & 0.26 & 0.52\\
        \midrule
        \multirow{4}{*}{\shortstack[l]{Divergent\\(n=34)}}
        & GPT-5.2 & 0.84 & 1.00 & 1.00\\
        & Sonnet-4.5 & 0.94 & 0.88 & 0.85\\
        & Gemini-3-Flash & 0.88 & 0.82 & 0.91\\
        & Gemini-3-Pro & 0.94 & 0.88 & 0.74\\
        \midrule
        \multirow{4}{*}{\shortstack[l]{Benign\\(n=35)}}
        & GPT-5.2 & 1.00 & 1.00 & 1.00\\
        & Sonnet-4.5 & 0.97 & 1.00 & 1.00\\
        & Gemini-3-Flash & 0.91 & 0.94 & 0.86\\
        & Gemini-3-Pro & 0.97 & 0.89 & 0.91\\
        \bottomrule
    \end{tabular}
\end{table}

\subsection{Ablation: Agentic Prompting Strategies}
Using the \tac{} dataset with GPT-5.2, we experiment across different prompting strategies like asking the agents to explicitly reason, plan or reflect. Full results are shown in Table~\ref{tab:agentic_prompting}.

\begin{table*}[ht]
    \centering
    \caption{Agentic Prompting Ablation (GPT-5.2 on TAC, n=85). Pass@3 and Ckpt\% show underspecified performance without ask\_user tool and values in parentheses shows improvement with the tool. \%Ask is percentage of trials where agent asked questions. Avg/Traj = mean number of questions per trajectory (among trajectories with user calls).}
    \label{tab:agentic_prompting}
    \vskip 0.1in
    \begin{small}
    \begin{tabular}{@{}llcccc@{}}
        \toprule
        \textbf{Class} & \textbf{Strategy} & \textbf{Pass@3} & \textbf{Ckpt\%} & \textbf{\%Ask} & \textbf{Avg/Traj} \\
        \midrule
        \multirow{4}{*}{\shortstack[l]{Overall\\(n=85)}}
        & CodeAct (Original) & \textbf{33.8} {\scriptsize(+6.5)} & \textbf{66.0} {\scriptsize(+2.7)} & 73\% & 1.8 \\
        & ReAct & 28.4 {\scriptsize(\textbf{+10.1})} & 62.5 {\scriptsize(+6.8)} & 69\% & 1.9 \\
        & Reflexion & 31.3 {\scriptsize(+8.7)} & 63.8 {\scriptsize(\textbf{+7.4})} & 68\% & 2.1 \\
        & Plan \& Execute & 31.3 {\scriptsize(+6.0)} & 62.7 {\scriptsize(+6.6)} & 80\% & 2.5 \\
        \midrule
        \multirow{4}{*}{\shortstack[l]{Outcome-\\Critical\\(n=40)}}
        & CodeAct (Original) & 5.0 {\scriptsize(+12.5)} & 52.8 {\scriptsize(+6.7)} & 74\% & 1.9 \\
        & ReAct & \textbf{5.3} {\scriptsize(+12.2)} & \textbf{53.1} {\scriptsize(+7.4)} & 68\% & 2.2 \\
        & Reflexion & 0.0 {\scriptsize(+17.9)} & 51.1 {\scriptsize(\textbf{+13.4})} & 68\% & 2.3 \\
        & Plan \& Execute & 0.0 {\scriptsize(\textbf{+18.4})} & 51.1 {\scriptsize(+12.0)} & 77\% & 2.5 \\
        \midrule
        \multirow{4}{*}{\shortstack[l]{Divergent\\(n=30)}}
        & CodeAct (Original) & \textbf{53.8} {\scriptsize($-$3.8)} & \textbf{75.5} {\scriptsize(+1.1)} & 73\% & 2.1 \\
        & ReAct & 44.0 {\scriptsize(\textbf{+7.9})} & 71.2 {\scriptsize(\textbf{+4.8})} & 74\% & 2.1 \\
        & Reflexion & 50.0 {\scriptsize(+2.0)} & 72.3 {\scriptsize(+2.0)} & 72\% & 2.2 \\
        & Plan \& Execute & 48.0 {\scriptsize($-$5.7)} & 69.0 {\scriptsize(+4.6)} & 79\% & 2.6 \\
        \midrule
        \multirow{4}{*}{\shortstack[l]{Benign\\(n=15)}}
        & CodeAct (Original) & \textbf{90.9} {\scriptsize(+9.1)} & \textbf{91.2} {\scriptsize($-$7.8)} & 58\% & 1.7 \\
        & ReAct & 72.7 {\scriptsize(+9.1)} & 74.7 {\scriptsize(\textbf{+9.3})} & 61\% & 1.8 \\
        & Reflexion & 100 {\scriptsize($-$9.1)} & 88.9 {\scriptsize($-$1.0)} & 55\% & 1.6 \\
        & Plan \& Execute & 100 {\scriptsize($-$9.1)} & 87.7 {\scriptsize($-$7.4)} & 88\% & 2.3 \\
        \bottomrule
    \end{tabular}
    \end{small}
\end{table*}

\subsection{Validating Outcome-Critical Candidate Generation}
We validate that our synthetic data pipeline can consistently generate outcome-critical ambiguities based on our filtering criteria. The results are shown in Table~\ref{tab:pipeline-results}.

\begin{table}[t]
\caption{Pipeline validation: severity strategies produce different outcome distributions on TAC (n=233 candidates, Gemini-3-Flash agent).}
\label{tab:pipeline-results}
\begin{center}
\begin{tabular}{@{}lccccc@{}}
\toprule
Strategy & Pass@3 & \shortstack{Outcome-\\Critical} & Divergent & Benign & \shortstack{New\\Task} \\
\midrule
\textsc{Delete} & 29.6\% & 146 & 54 & 15 & 18 \\
\textsc{Vaguify} & 19.2\% & 173 & 39 & 5 & 12 \\
\textsc{Genericize} & 27.5\% & 151 & 54 & 10 & 18 \\
\bottomrule
\end{tabular}

\end{center}
\end{table}

\subsection{Critical Blocker Property}
In Table \ref{tab:critical_blocker}, we show the performance on the outcome-critical subset, where agents have near-complete failure without clarification.

\begin{table}[ht]
\caption{Fidelity: Outcome-critical variants Pass@3 (\%) show near-complete failure without clarification.}
\label{tab:critical_blocker}
\vskip 0.1in
\begin{center}
\begin{small}
\begin{tabular}{@{}llccc@{}}
\toprule
\textbf{Dataset} & \textbf{Model} & \textbf{Base} & \textbf{Under.} & \textbf{+User} \\
\midrule
\multirow{5}{*}{\shortstack[l]{MCP-Atlas\\(n=50)}}
& Opus-4.5 & 1.00 & 0.14 & 0.70 \\
& Sonnet-4.5 & 0.87 & 0.06 & 0.58 \\
& Gemini-3-Pro & 1.00 & 0.08 & 0.62 \\
& Gemini-3-Flash & 0.93 & 0.10 & 0.42 \\
& GPT-5.2 & 1.00 & 0.14 & 0.74 \\
\midrule
\multirow{5}{*}{\shortstack[l]{TAC\\(n=40)}}
& Opus-4.5 & 0.54 & 0.07 & 0.20 \\
& Sonnet-4.5 & 0.45 & 0.02 & 0.15 \\
& Gemini-3-Pro & 0.45 & 0.15 & 0.18 \\
& Gemini-3-Flash & 0.63 & 0.07 & 0.19 \\
& GPT-5.2 & 0.54 & 0.05 & 0.17 \\
\midrule
\multirow{4}{*}{\shortstack[l]{SWE-Bench\\Pro (n=31)}}
& GPT-5.2 & 0.72 & 0.00 & 0.71 \\
& Sonnet-4.5 & 0.96 & 0.45 & 0.48 \\
& Gemini-3-Flash & 0.80 & 0.39 & 0.35 \\
& Gemini-3-Pro & 0.84 & 0.26 & 0.52 \\
\bottomrule
\end{tabular}
\end{small}
\end{center}
\vskip -0.1in
\end{table}

\subsection{Cross-Benchmark Generalization}
In Table~\ref{tab:generality}, we include aggregated Pass@3 (\%) across models for each benchmark.

\begin{table}[ht]
\caption{Generality: Aggregated Pass@3 (\%) across benchmarks (model average).}
\label{tab:generality}
\vskip 0.1in
\begin{center}
\begin{small}
\begin{tabular}{@{}lccc@{}}
\toprule
\textbf{Dataset} & \textbf{Base} & \textbf{Under.} & \textbf{+User} \\
\midrule
MCP-Atlas (n=100) & 0.96 & 0.44 & 0.73 \\
TAC (n=85) & 0.66 & 0.30 & 0.37 \\
SWE-Bench Pro (n=100) & 0.91 & 0.73 & 0.79 \\
\bottomrule
\end{tabular}
\end{small}
\end{center}
\vskip -0.1in
\end{table}

\FloatBarrier
\section{Benchmark Dataset Distribution}
\label{app:dataset-distribution}

\begin{figure*}[!ht]
\begin{center}
\centerline{\includegraphics[width=\textwidth]{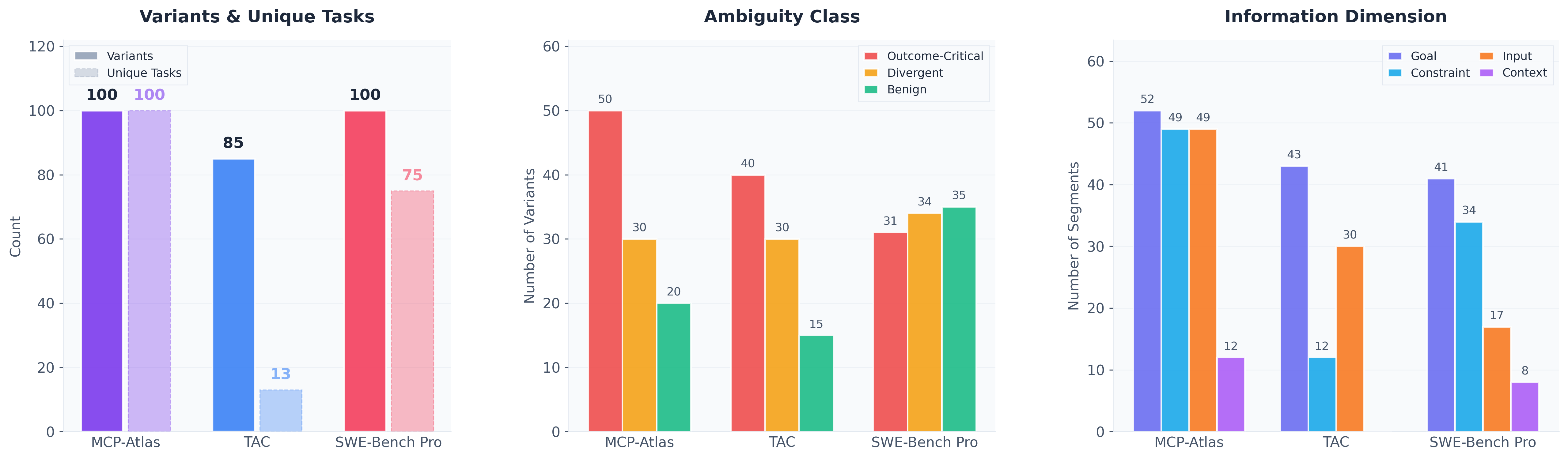}}
\caption{\textbf{LHAW Benchmark Distribution.} Final dataset of 285 variants across three benchmarks. \textit{Left:} variant and unique task counts per dataset. \textit{Center:} ambiguity class distribution (outcome-critical, divergent, benign). \textit{Right:} information dimension distribution (goal, constraint, input, context); multi-segment removals in MCP-Atlas count each dimension separately.}
\label{fig:dataset_distribution}
\end{center}
\vskip -0.1in
\end{figure*}

Figure~\ref{fig:dataset_distribution} summarizes the final \lhaw{} benchmark of 285 underspecified variants. The left panel shows variant and unique task counts per dataset, highlighting the 1:1 variant-to-task mapping in \mcpatlas{} versus the many-to-one mappings in \tac{} (85 variants from 13 tasks) and \swebench{} (100 variants from 75 tasks). The center panel shows the ambiguity class distribution: \mcpatlas{} skews toward outcome-critical variants (50\%), while \swebench{} is more balanced across classes. The right panel shows the information dimension distribution, where multi-segment removals in \mcpatlas{} are counted per segment, yielding 162 total dimension instances from 100 variants. Goal is the most prevalent dimension overall, while TAC contains no context-type underspecification.

\FloatBarrier
\section{Synthetic Pipeline Implementation Details}
\label{app:pipeline}

\subsection{Segment Extraction Prompt}
\label{app:extraction-prompt}

The segment extraction prompt instructs the LLM to identify removable information segments with structured scoring.

\begin{lstlisting}
You are an expert at analyzing task prompts to identify 
critical information segments.

A SEGMENT is an atomic piece of information that, if removed, 
would create outcome-critical ambiguity.

## Taxonomy (4 dimensions)
1. GOAL       - WHAT to produce (output file, format, criteria)
2. CONSTRAINT - HOW to do it (thresholds, rules, methods)
3. INPUT      - FROM WHERE (file paths, data sources)
4. CONTEXT    - WHAT BACKGROUND (jargon, business rules)

## Segment Fields
- text: Short verbatim span with context around key info
- value: The ATOMIC piece of info (filename, number, rule)
         Must be exact substring from prompt, keep SHORT
- dimension / subdimension: From taxonomy above

## Scoring (use ONLY 0.0, 0.5, or 1.0)

criticality - What happens if agent uses WRONG value?
  1.0: Task FAILS   (wrong output file created)
  0.5: Task PARTIAL (degraded result)
  0.0: Task OK      (cosmetic difference only)

guessability - If removed, can agent recover correct value?
  0.0: Impossible   (specific output filename - must be told)
  0.5: Discoverable (input file - could run `ls` to find)
  1.0: Obvious      (universal default or stated elsewhere)

## Grounding (if trajectory/checkpoints provided)
- is_used_in_trajectory: bool
- first_use_pct: float (0.0=start, 1.0=end)
- checkpoint_refs: ["CP1", "CP3"]

## Output (JSON)
{
  "segments": [
    {
      "text": "<verbatim span>",
      "value": "<shortest key info>",
      "dimension": "input|goal|constraint|context",
      "subdimension": "<from taxonomy>",
      "criticality": 1.0,
      "guessability": 0.5,
      "is_used_in_trajectory": true,
      "first_use_pct": 0.15,
      "checkpoint_refs": ["CP1"]
    }
  ]
}
\end{lstlisting}

\subsection{Variant Generation Prompt}
\label{app:generation-prompt}

The variant generation prompt creates underspecified versions by removing or obscuring segments.

\begin{lstlisting}
You are creating an underspecified version of a task prompt.

Goal: Create outcome-critical ambiguity where the agent could 
reasonably interpret the prompt in multiple ways leading to 
different task outcomes.

## Segment to Remove
- Dimension: {dimension}
- Value: "{value}"
- Text: "{text}"

## Removal Strategy
{severity_instruction}

## Instructions
1. Remove/obscure the segment - including ALL occurrences
2. The removed info should be genuinely missing - agent could 
   assume WRONG value resulting in task failure
3. Keep result grammatically correct and natural-sounding
4. Preserve exact formatting (line breaks, whitespace, markdown)

## Output (JSON)
{
  "underspecified_prompt": "...",
  "expected_questions": [{"segment_id": "S1", "questions": [...]}],
  "failure_mode": "What goes wrong if agent assumes wrong value"
}
\end{lstlisting}

\subsection{Severity Instructions}

The \texttt{\{severity\_instruction\}} placeholder is filled with one of:

\begin{lstlisting}
DELETE:    "Remove this information entirely, adjusting 
                   sentence structure to remain natural."

VAGUIFY: "Replace with vague language 
                   (e.g., 'the file', 'the output')."

GENERICIZE: "Replace with generic reference 
                   (e.g., 'appropriate format', 'suitable location')."
\end{lstlisting}

\subsection{New Task Judge Prompt}
\label{appendix:new-task-judge-prompt}
During the empirical agent trials process, we also apply a filtering step to remove underspecified variants that have been transformed into a new but valid task. To remove these samples, we leverage an LLM judge (Gemini-3-Flash) to determine new task eligibility with the prompt:

\begin{lstlisting}
You are an impartial judge evaluating whether an AI assistant has successfully completed a task.

## Task Given to Assistant
{underspec_prompt}

## Assistant's Final Responses (All Trials)
{formatted_responses}

## Your Evaluation Task
Determine whether ALL trials reached the SAME answer/conclusion AND that answer completes the task AS SPECIFIED in the prompt above.

Note: the model should *NOT* make any assumptions about the task beyond what is explicitly stated in the prompt. If you see it making any connections that are not an explicit ask in the prompt then it should call task_completed as false.

## Response Format
Provide your evaluation in the following JSON format:
{{
    "task_completed": true/false,
    "reasoning": "Brief explanation of your verdict"
}}

Respond ONLY with the JSON object, no other text.
\end{lstlisting}

\FloatBarrier
\subsection{TAC Original Task Breakdown}
In Table~\ref{tab:tasks}, we show the filtered task breakdown from OwnCloud after running initial pass@3 baselines and computing partial task completion.

\begin{table}[t]
\caption{Selected TAC tasks by partial completion (checkpoint accuracy $\geq$50\%). ``Solvable'' indicates \passmark{} $> 0$ for any reference models we use for partial task completion filtering.}
\label{tab:tasks}
\vskip 0.1in
\begin{center}
\begin{small}
\begin{tabular}{@{}lccc@{}}
\toprule
Domain & Tasks & Avg Ckpt & Solvable \\
\midrule
Data Science & 5 & 72\% & 4/5 \\
Finance & 3 & 75\% & 1/3 \\
HR & 4 & 81\% & 3/4 \\
SDE & 1 & 78\% & 1/1 \\
\midrule
\textbf{Total} & \textbf{13} & \textbf{76\%} & \textbf{9/13} \\
\bottomrule
\end{tabular}
\end{small}
\end{center}
\vskip -0.1in
\end{table}



\subsection{Example of Original vs Underspecified Prompt}
Table~\ref{tab:removal_example} shows how the three removal strategies transform a task from \tac{} (ds-format-excel-sheets) when removing two constraint segments: the color code ``\#87CEEB'' and the alignment ``horizontally centered''. The original segments are highlighted in green, while the vaguified/genericized replacements are highlighted in blue.

\begin{table*}[ht]
\caption{Removal strategies applied to ds-format-excel-sheets (removed segments: \textbf{\#87CEEB}, \textbf{horizontally centered}). Each row shows the full prompt under a different strategy. Highlighted text indicates modified segments.}
\label{tab:removal_example}
\vskip 0.1in
\begin{center}
\resizebox{\textwidth}{!}{%
\begin{tabular}{@{}p{2.0cm}p{15cm}@{}}
\toprule
\textbf{Strategy} & \textbf{Prompt} \\
\midrule
\textbf{Original} &
Find the \texttt{respondent\_counts.xlsx} file under \texttt{Documents/Data Analysis} on Owncloud, and based on the source ``unformatted'' sheet, create a ``formatted'' sheet and make formatting edits based on the following requirements:\newline
1. Change the background color of top-header cells to skyblue (\colorbox{green!20}{\textbf{\#87CEEB}})\newline
2. Make sure all cell values are \colorbox{green!20}{\textbf{horizontally centered}} in their cells\newline
Do not change the ``unformatted'' sheet, only make changes in the ``formatted'' sheet. Save all your changes to \texttt{respondent\_counts.xlsx} on owncloud. \\
\addlinespace
\textsc{Delete} &
Find the \texttt{respondent\_counts.xlsx} file under \texttt{Documents/Data Analysis} on Owncloud, and based on the source ``unformatted'' sheet, create a ``formatted'' sheet and make formatting edits based on the following requirements:\newline
1. Change the background color of top-header cells\newline
2. Make sure all cell values are in their cells\newline
Do not change the ``unformatted'' sheet, only make changes in the ``formatted'' sheet. Save all your changes to \texttt{respondent\_counts.xlsx} on owncloud. \\
\addlinespace
\textsc{Vaguify} &
Find the \texttt{respondent\_counts.xlsx} file under \texttt{Documents/Data Analysis} on Owncloud, and based on the source ``unformatted'' sheet, create a ``formatted'' sheet and make formatting edits based on the following requirements:\newline
1. Change the background color of top-header cells to \colorbox{blue!15}{a specific shade of blue}\newline
2. Make sure all cell values are \colorbox{blue!15}{centered} in their cells\newline
Do not change the ``unformatted'' sheet, only make changes in the ``formatted'' sheet. Save all your changes to \texttt{respondent\_counts.xlsx} on owncloud. \\
\addlinespace
\textsc{Genericize} &
Find the \texttt{respondent\_counts.xlsx} file under \texttt{Documents/Data Analysis} on Owncloud, and based on the source ``unformatted'' sheet, create a ``formatted'' sheet and make formatting edits based on the following requirements:\newline
1. Change the background color of top-header cells to \colorbox{blue!15}{an appropriate color}\newline
2. Make sure all cell values are \colorbox{blue!15}{suitably aligned} in their cells\newline
Do not change the ``unformatted'' sheet, only make changes in the ``formatted'' sheet. Save all your changes to \texttt{respondent\_counts.xlsx} on owncloud. \\
\bottomrule
\end{tabular}%
}
\end{center}
\vskip -0.1in
\end{table*}

\FloatBarrier
\subsection{Example Underspecified Goal Variant from \tac{}}
\label{app:examples}

\paragraph{Original:}
\begin{quote}
\small
Navigate to ownCloud at ``/Documents/HR/Attendance''. Use ``april-attendance-data.csv'' and ``salary-rates.pdf'' to \textbf{calculate the total earnings for each employee}. Create ``april-payroll.xlsx'' with columns `Name' and `Total Earnings'.
\end{quote}

\paragraph{Underspecified:}
\begin{quote}
\small
Navigate to ownCloud at ``/Documents/HR/Attendance''. Use ``april-attendance-data.csv'' and ``salary-rates.pdf''. Create ``april-payroll.xlsx'' with columns `Name' and `Total Earnings'.
\end{quote}

\paragraph{Expected Question:}
``What calculation should I perform with the attendance and salary data?''

\FloatBarrier
\section{Underspecification Taxonomy: Subdimensions}
\label{app:taxonomy}

Each of the four information dimensions from Table~\ref{tab:taxonomy} has finer subdimensions for precise categorization of removed segments:

\paragraph{Goal Subdimensions}
\begin{itemize}[noitemsep]
    \item \textbf{target}: The specific thing to produce, fix, or accomplish (e.g., ``report.xlsx'', ``the login bug'')
    \item \textbf{format}: File type, output structure, or response format (e.g., ``xlsx'', ``JSON'', ``markdown table'')
    \item \textbf{content}: Required fields, sections, or elements in output (e.g., ``Name and Amount columns'')
    \item \textbf{acceptance}: Criteria defining task success (e.g., ``tests pass'', ``matches reference'')
\end{itemize}

\paragraph{Constraint Subdimensions}
\begin{itemize}[noitemsep]
    \item \textbf{numeric\_bound}: Thresholds, limits, counts, or quantities (e.g., ``top 5'', ``under \$500'')
    \item \textbf{precision}: Rounding, truncation, significant figures (e.g., ``2 decimal places'')
    \item \textbf{temporal}: Time ranges, deadlines, periods (e.g., ``last 12 months'', ``Q1 2024'')
    \item \textbf{selection}: Criteria for including/excluding items (e.g., ``only active'', ``where status=open'')
    \item \textbf{method}: Algorithm, approach, or technique (e.g., ``sort by date desc'', ``apply FIFO'')
\end{itemize}

\paragraph{Input Subdimensions}
\begin{itemize}[noitemsep]
    \item \textbf{source}: Where to get data---file, API, database, service (e.g., ``from sales.csv'', ``via Stripe API'')
    \item \textbf{location}: Path, URL, endpoint, or address (e.g., ``/data/reports/'', ``s3://bucket/key'')
    \item \textbf{identifier}: Specific name, ID, or reference (e.g., ``order \#12345'', ``ticket JIRA-567'')
    \item \textbf{version}: Which version, snapshot, or state (e.g., ``v2.0'', ``latest stable'')
\end{itemize}

\paragraph{Context Subdimensions}
\begin{itemize}[noitemsep]
    \item \textbf{terminology}: Domain-specific terms, acronyms (e.g., ``MRR'', ``P0 bug'', ``qualified lead'')
    \item \textbf{business\_logic}: Organization-specific rules or calculations (e.g., ``overtime = 1.5x after 40h'')
    \item \textbf{conventions}: Implicit standards or norms (e.g., ``dates in ISO format'', ``amounts in USD'')
    \item \textbf{tool\_knowledge}: How to use specific tools or systems (e.g., ``git workflow'', ``JIRA board structure'')
\end{itemize}

\FloatBarrier
\section{Data Structure}
\label{app:data-structure}

After the synthetic pipeline processes a task, the output evolves through two phases:

\subsection{Phase 2 Output (Post-Generation)}
After candidate generation, each variant includes full segment metadata for filtering and analysis:

\begin{lstlisting}
{
  "task_id": "finance_check_attendance_payroll_V_S3",
  "agent_prompt": "Create payroll using the attendance file...",
  "removed_segments": [
    {
      "id": "S1",
      "dimension": "input",
      "subdimension": "identifier",
      "value": "april-attendance-data.csv",
      "text": "april-attendance-data.csv",
      // Grounding (LLM analyzes trajectory + checkpoints)
      "is_used_in_trajectory": true,
      "first_use_pct": 0.15,
      "checkpoint_refs": ["CP1", "CP2"],
      // Scores (LLM self-assesses, 3-level scale)
      "criticality": 1.0,   // 0.0 (OK) | 0.5 (PARTIAL) | 1.0 (FAILS)
      "guessability": 0.5,  // 0.0 (cannot) | 0.5 (maybe) | 1.0 (will)
      "priority_score": 0.5 // criticality * (1 - guessability)
    }
  ],
  "criteria": {
    "severity": "delete"  // DELETE | VAGUIFY | GENERICIZE
  },
  "expected_questions": [
    {"S1": ["Which attendance file should I use?", ...]}
  ],
  "expected_failure_mode": "Agent uses wrong file, produces incorrect payroll"
}
\end{lstlisting}

\subsection{Phase 3 Output (Benchmark-Ready)}
After empirical agent trials, variants are classified and formatted for public release:

\begin{lstlisting}
[
  {
    "variant_id": "hr_new_grad_job_description_V10_goal",
    "underspecified_prompt": "Write a job description...",
    "information_dimension": ["goal"],  // goal | constraint | input | context
    "ambiguity_class": "benign",      // outcome-critical | divergent | benign
    "removed_segments": [
      {
        "id": "S1",
        "dimension": "goal",
        "subdimension": "target",
        "value": "new grad software engineer job description",
      }
    ],
    "expected_questions": [
      {
        "segment_id": "S1",
        "questions": ["What type of job description should I create?"]
      }
    ],
    "terminal_states": "[(1, 1)]"
    "original_prompt": "Write a new grad software engineer job...",
    "original_task": "hr_new_grad_job_description",
    "dataset": "TheAgentCompany",
  },
  // ... additional samples
]
\end{lstlisting}

\subsection{Field Descriptions}
\begin{itemize}
    \item \textbf{variant\_id}: Unique identifier encoding task name, version, and removed dimensions
    \item \textbf{original\_task}: Base task name from TheAgentCompany
    \item \textbf{original\_prompt}: Fully-specified prompt with all information
    \item \textbf{underspecified\_prompt}: Modified prompt with critical information removed
    \item \textbf{information\_dimension}: Primary type of missing information (goal, constraint, input)
    \item \textbf{ambiguity\_class}: Empirical classification based on pass@k results
    \item \textbf{removed\_segments}: List of information segments that were removed, each with dimension, subdimension (e.g., target, temporal, location), and the exact value
    \item \textbf{expected\_questions}: Clarifying questions an ideal agent should ask
    \item \textbf{terminal\_states}: Unique checkpoint pass/fail tuples from Pass@K (k=3) evaluation runs.
\end{itemize}

\section{User Simulator Tool Implementation}
\label{app:user-tool}

We implement the user simulator as an MCP (Model Context Protocol) server, enabling agents to invoke it via a standardized tool interface. This design allows seamless integration with any MCP-compatible agent framework without modifying the agent's core implementation.

\subsection{Tool Interface}

The agent interacts with the user simulator through the \texttt{ask\_user} tool:

\begin{lstlisting}
@mcp.tool()
async def ask_user(
    question: str,
    context: str = "",
) -> str:
    """Ask the user a clarifying question to get more 
    information about the task. Use this when the task 
    is ambiguous or you need specific details to proceed.

    Args:
        question: The clarifying question to ask the user.
        context: Optional additional context (e.g., 
                 conversation history summary)

    Returns:
        The user's response with the requested information.
    """
\end{lstlisting}

\subsection{User Simulator Prompt}

The user simulator is an LLM with access to: (1) the original well-specified task, (2) the underspecified prompt shown to the agent, and (3) the list of removed segment values. Below is the system prompt:

\begin{lstlisting}
You are simulating a user who has a task in mind but 
didn't fully specify it.

The user originally intended to give this COMPLETE prompt:
{primary_task}

But they actually gave this UNDERSPECIFIED version:
{underspecified_prompt}

The parts that were removed/made vague:
{removed_values}

An AI assistant (who only sees the underspecified version) 
is now asking you a clarifying question.

Your job: Compare the two prompts, find what's MISSING from 
the underspecified version, and provide the EXACT information 
from the complete prompt.

Guidelines:
- Find the EXACT values that are in the complete prompt but 
  missing from the underspecified one
- Provide those specific values (times, names, dates, 
  numbers, phrases, etc.)
- Be concise - just answer what's asked
- Don't reveal you're a simulation

{additional_context}
\end{lstlisting}

The user simulator uses GPT-4.1 by default (\texttt{USER\_SIMULATOR\_MODEL} environment variable) with temperature 0.7 to provide natural variation in response style while maintaining factual accuracy.

\subsection{User Simulator Prompt for \mcpatlas{}}

In order to get the user to answer the agent in a more natural conversation flow for \mcpatlas{}, we had to slightly adjust the user prompt to encourage it to be realistic, but not reveal more (e.g., ``say something like you aren't sure or can't remember''), while also ensuring it didn't hallucinate information. The following prompt worked well in our trials:

\begin{lstlisting}
**System Role:** You are simulating a human user. You are currently in a conversation with an AI Agent who is trying to help you with a task. 

### THE SITUATION:
You have an **Original Well-Scoped Task** which represents the set of instructions you intended to provide and EVERYTHING you know about the task. However, the message you actually sent to the Agent (**Your Sent Message**) was an underspecified version where some details were removed or made vague.

The Agent is now asking you for more details, which you may or may not be able to help with.

---

### YOUR INTERNAL CONTEXT:
* **Original Well-Scoped Task (What you intended):** > {original_prompt}

* **Your Sent Message (What the Agent saw):** > {underspecified_prompt}

* **The Specific Details You Left Out:** > {removed_values}

---

### KNOWLEDGE & LOGIC BOUNDARIES:

1. **The Knowledge Limit:** You LITERALLY CAN ONLY provide any information that is explicitly stated in the **Original Well-Scoped Task**. You CANNOT answer any questions where the answer is not in the Original Well-Scoped Task.
2. **Strict Data Retrieval:**
    * If the Agent asks for a detail that is **NOT** explicitly written in the Original Well-Scoped Task, you do not know it. 
    * Do not "figure it out," guess, or use common sense to fill in gaps that aren't in your original notes.
3. **No Hand-Holding:** The Agent has its own tools and intelligence that can help it figure things out (which you can't see). Do not explain "how" to do the task or provide extra context. Just give the information that is in the Original Well-Scoped Task but not in the Sent Message.
4. **Persona:** Be a concise, direct human user. 
5. **Compound Questions:** If the agent asks for multiple details at once, and you only have information for **some** of them:
    - Provide the exact information you **do** have from the Original Well-Scoped Task.
    - For the missing parts, say something like you aren't sure or can't remember. 


REMEMBER DO NOT GUESS OR MAKE UP ANYTHING THAT IS NOT IN THE ORIGINAL WELL-SCOPED TASK.
\end{lstlisting}

\subsection{Agent System Prompt}
\label{app:agent-prompt}

To inform agents that clarification is available, we append the following to the agent's system prompt:

\begin{lstlisting}
IMPORTANT: Your output will be checked by an auto-grader 
looking for exact answers.
This task may be missing critical information.
Use the ask_user tool to ask the user for any missing details.
\end{lstlisting}

This prompt serves two purposes: (1) it signals that precision matters (discouraging arbitrary assumptions), and (2) it makes the agent aware that the \texttt{ask\_user} tool is available. Importantly, it does not \emph{require} the agent to ask questions---the agent must still decide whether clarification is necessary based on its assessment of missing information.

\section{Evaluation Harness Implementation}
\label{app:eval-harness}

We provide details on how we adapted each benchmark's evaluation infrastructure to support our underspecification experiments.

\paragraph{TheAgentCompany (TAC).}
For \tac{} experiments, we adapt the OpenHands-based evaluation harness from \citet{kale2025mrt}, which provides a parallel task execution framework for running agents on enterprise workflow tasks. We copy the official evaluation files (task definitions, checkpoint specifications, and grading scripts) from the \tac{} repository~\cite{theagentcompany2024} into our codebase, ensuring evaluation fidelity with the original benchmark. The harness orchestrates Docker-based agent execution with configurable parallelism and integrates our ask\_user MCP tool for user simulation experiments.

\paragraph{SWE-Bench Pro.}
For \swebench{} experiments, we use the official evaluation infrastructure from the SWE-Bench Pro repository.\footnote{\url{https://github.com/scaleapi/SWE-bench_Pro-os}} The harness uses Modal for distributed evaluation and pre-built Docker images for reproducible instance environments. We generate patches using the SWE-agent scaffold and evaluate using the provided \texttt{swe\_bench\_pro\_eval.py} script, which applies patches to the target repositories and runs the test suites to determine pass/fail outcomes.

\paragraph{MCP-Atlas.}
For \mcpatlas{} experiments, we follow the official evaluation setup from the benchmark repository~\cite{bandi2026mcpatlaslargescalebenchmarktooluse}. We configure API keys for each MCP server integration and run agents through the provided harness, which scores task completion based on checkpoint criteria. Each task requires the agent to invoke appropriate MCP tools to complete the specified objectives.

\paragraph{Model Configuration.}
All experiments use official API endpoints for frontier models (Claude Opus-4.5, Claude Sonnet-4.5, Gemini-3-Pro, Gemini-3-Flash, GPT-5.2) through LiteLLM. We run 3 independent trials per variant to compute pass@3 metrics.

\section{Value of Information: Additional Details} \label{sec:value_of_info_more}

\begin{figure}[h]
\begin{center}
\centerline{\includegraphics[width=\columnwidth]{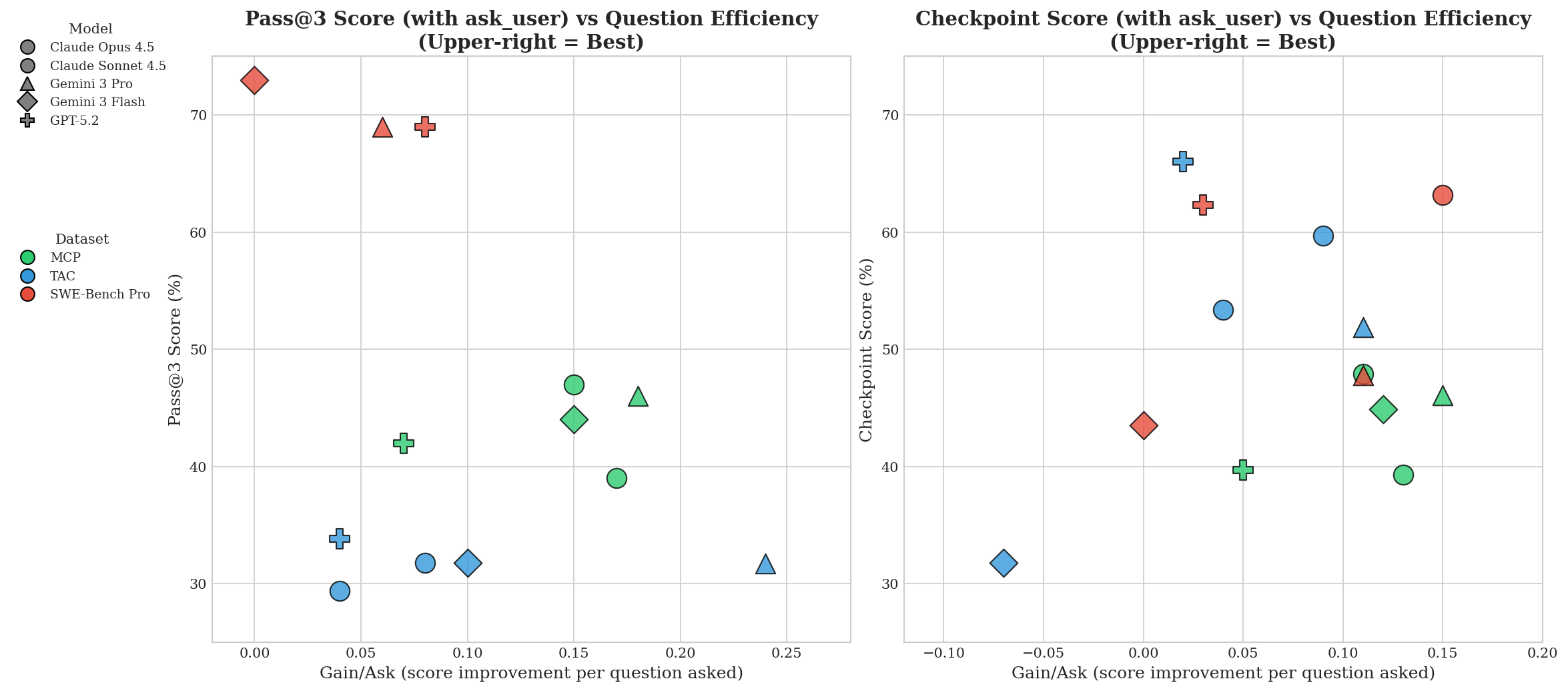}}
\caption{\textbf{Value of Information} across models across tasks. This shows the pass@3 and Avg. Ckpt\% overall performance with the ask\_user tool plotted against the gain provided by each user question. The top right is the most capable agents who learn the most per user call. The bottom left is the least capable agents who learn the least per user call.}
\label{fig:agent_efficiency}
\end{center}
\vskip -0.1in
\end{figure}

In Figure~\ref{fig:agent_efficiency}, we see the relationship between overall task performance and clarification efficiency across models. Models in the upper-right quadrant achieve both high task success and high value per question asked, indicating effective clarification behavior.

\FloatBarrier
\section{Ask User Analysis Details}\label{sec:ask_user}

\subsection{Prompt for Judging Ask User into Taxonomy} \label{sec:ask-user-judge}

Below is the prompt fed to the Judge:
\begin{lstlisting}

Failure Analysis Categories:
    1. Question Quality Deficiency: The agent asks questions, but they are poorly formulated and fail to elicit the necessary information.
    2. Question Targeting Error: The agent asks questions about the wrong aspects of the task, missing the critical missing information.
    3. Information Integration Failure: The agent receives correct information but fails to properly incorporate it into task execution.
    4. Over-Clarification: The agent asks excessive questions, including about information that was not removed or could be inferred.
    5. Under-Clarification: The agent asks too few questions relative to the amount of critical missing information.
    6. Timing & Strategy Error: The agent asks questions at inappropriate times or in an inefficient sequence.
    7. User Response Misinterpretation: The agent misunderstands or incorrectly processes the user's clarifying responses.

Detailed Analysis Flags:
For each flag below, choose YES, NO, or N/A based on the strict definitions provided:

1. Question Quality Deficiencies
  - `flag_vague_questions`:
    - Definition: Whether the agent's questions are too vague or ambiguous to elicit specific, actionable information.
    - Prerequisite: The agent asked at least one clarification question.
    - `YES`: Questions lack specificity needed to recover precise information (e.g., "What file?" instead of "Which attendance file should I use - april-attendance.csv or may-attendance.csv?").
    - `NO`: The prerequisite IS met, but the conditions for `YES` are NOT satisfied.
    - `N/A`: The prerequisite is NOT met.

  - `flag_compound_questions`:
    - Definition: Whether the agent bundles multiple distinct questions together, making it difficult for users to provide complete answers.
    - Prerequisite: The agent asked at least one clarification question.
    - `YES`: Questions contain multiple sub-questions that should be asked separately.
    - `NO`: The prerequisite IS met, but the conditions for `YES` are NOT satisfied.
    - `N/A`: The prerequisite is NOT met.

  - `flag_closed_questions`:
    - Definition: Whether the agent asks yes/no or multiple choice questions when open-ended questions are needed.
    - Prerequisite: The removed segment requires complex or detailed information.
    - `YES`: Agent uses closed questions for information requiring detailed responses.
    - `NO`: The prerequisite IS met, but the conditions for `YES` are NOT satisfied.
    - `N/A`: The prerequisite is NOT met.

2. Question Targeting Errors
  - `flag_missed_critical_segments`:
    - Definition: Whether the agent fails to ask about outcome-critical missing information.
    - Prerequisite: At least one removed segment has criticality >= 0.5 and guessability <= 0.5.
    - `YES`: Agent does not ask about critical segments that required clarification.
    - `NO`: The prerequisite IS met, but the conditions for `YES` are NOT satisfied.
    - `N/A`: The prerequisite is NOT met.

  - `flag_wrong_dimension_focus`:
    - Definition: Whether the agent asks about the wrong dimension of missing information.
    - Prerequisite: Multiple dimensions have removed segments.
    - `YES`: Agent focuses questions on dimensions with low criticality while ignoring high criticality dimensions.
    - `NO`: The prerequisite IS met, but the conditions for `YES` are NOT satisfied.
    - `N/A`: The prerequisite is NOT met.

  - `flag_inferable_information_asked`:
    - Definition: Whether the agent asks about information that could be inferred from the environment or context.
    - Prerequisite: At least one removed segment has guessability >= 0.5.
    - `YES`: Agent asks about information marked as inferable that it could have discovered through exploration.
    - `NO`: The prerequisite IS met, but the conditions for `YES` are NOT satisfied.
    - `N/A`: The prerequisite is NOT met.

3. Information Integration Failures
  - `flag_ignoring_user_response`:
    - Definition: Whether the agent receives correct information but does not use it in subsequent actions.
    - Prerequisite: The agent asked at least one question AND received a correct response.
    - `YES`: Agent's actions after receiving information do not reflect the provided details.
    - `NO`: The prerequisite IS met, but the conditions for `YES` are NOT satisfied.
    - `N/A`: The prerequisite is NOT met.

  - `flag_incorrect_information_application`:
    - Definition: Whether the agent applies the received information incorrectly to the task.
    - Prerequisite: The agent received correct information from the user.
    - `YES`: Agent uses the information but applies it to the wrong part of the task or in the wrong way.
    - `NO`: The prerequisite IS met, but the conditions for `YES` are NOT satisfied.
    - `N/A`: The prerequisite is NOT met.

  - `flag_partial_information_utilization`:
    - Definition: Whether the agent only uses part of the information provided in user responses.
    - Prerequisite: User responses contain multiple pieces of information.
    - `YES`: Agent extracts and uses only a subset of the information provided.
    - `NO`: The prerequisite IS met, but the conditions for `YES` are NOT satisfied.
    - `N/A`: The prerequisite is NOT met.

4. Over-Clarification Issues
  - `flag_redundant_questions`:
    - Definition: Whether the agent asks the same or highly similar questions multiple times.
    - Prerequisite: The agent asked multiple questions.
    - `YES`: Agent repeats questions or asks about information already provided.
    - `NO`: The prerequisite IS met, but the conditions for `YES` are NOT satisfied.
    - `N/A`: The prerequisite is NOT met.

  - `flag_unnecessary_questions`:
    - Definition: Whether the agent asks about information that was not removed or is clearly stated in the task.
    - Prerequisite: None.
    - `YES`: Agent asks questions about information already present in the task prompt.
    - `NO`: The conditions for `YES` are NOT satisfied.

  - `flag_excessive_cost`:
    - Definition: Whether the number of questions asked is disproportionate to the performance improvement achieved.
    - Prerequisite: The agent asked at least 3 questions.
    - `YES`: Cost-efficiency is low (< 0.3) despite multiple questions being asked.
    - `NO`: The prerequisite IS met, but the conditions for `YES` are NOT satisfied.
    - `N/A`: The prerequisite is NOT met.

5. Under-Clarification Issues
  - `flag_premature_execution`:
    - Definition: Whether the agent proceeds with task execution before clarifying critical missing information.
    - Prerequisite: Multiple segments were removed with high criticality.
    - `YES`: Agent begins critical task actions before asking necessary questions.
    - `NO`: The prerequisite IS met, but the conditions for `YES` are NOT satisfied.
    - `N/A`: The prerequisite is NOT met.

  - `flag_insufficient_question_depth`:
    - Definition: Whether the agent asks too few questions relative to the amount of missing critical information.
    - Prerequisite: Multiple critical segments were removed.
    - `YES`: Information recovery rate < 0.5 and ask_recall < 0.5.
    - `NO`: The prerequisite IS met, but the conditions for `YES` are NOT satisfied.
    - `N/A`: The prerequisite is NOT met.

  - `flag_false_confidence`:
    - Definition: Whether the agent proceeds with assumptions instead of clarifying when uncertainty is warranted.
    - Prerequisite: The agent makes explicit or implicit assumptions about missing information.
    - `YES`: Agent states or demonstrates assumptions without seeking validation.
    - `NO`: The prerequisite IS met, but the conditions for `YES` are NOT satisfied.
    - `N/A`: The prerequisite is NOT met.

6. Timing & Strategy Errors
  - `flag_late_clarification`:
    - Definition: Whether the agent asks questions too late in the execution, after the information was needed.
    - Prerequisite: The agent asked at least one question AND information is used early in task execution.
    - `YES`: Questions are asked after the point where the information would have been useful.
    - `NO`: The prerequisite IS met, but the conditions for `YES` are NOT satisfied.
    - `N/A`: The prerequisite is NOT met.

  - `flag_upfront_question_dump`:
    - Definition: Whether the agent asks all questions at the beginning without attempting any exploration.
    - Prerequisite: The agent asked multiple questions (>=3).
    - `YES`: All questions are asked in the first 10% of the trajectory without intermediate exploration.
    - `NO`: The prerequisite IS met, but the conditions for `YES` are NOT satisfied.
    - `N/A`: The prerequisite is NOT met.

  - `flag_inefficient_question_sequence`:
    - Definition: Whether the order of questions is illogical or inefficient for task progression.
    - Prerequisite: The agent asked multiple questions (>=2).
    - `YES`: Questions are asked in an order that prevents building on previous answers or requires backtracking.
    - `NO`: The prerequisite IS met, but the conditions for `YES` are NOT satisfied.
    - `N/A`: The prerequisite is NOT met.

7. User Response Misinterpretation
  - `flag_literal_misunderstanding`:
    - Definition: Whether the agent misinterprets the literal meaning of user responses.
    - Prerequisite: User provided clear, direct responses.
    - `YES`: Agent demonstrates misunderstanding of explicit information in responses.
    - `NO`: The prerequisite IS met, but the conditions for `YES` are NOT satisfied.
    - `N/A`: The prerequisite is NOT met.

  - `flag_context_misapplication`:
    - Definition: Whether the agent fails to contextualize user responses within the broader task.
    - Prerequisite: User responses require task context to be properly applied.
    - `YES`: Agent uses information out of context or misapplies it to wrong task components.
    - `NO`: The prerequisite IS met, but the conditions for `YES` are NOT satisfied.
    - `N/A`: The prerequisite is NOT met.

  - `flag_implicit_information_missed`:
    - Definition: Whether the agent misses implicit information or implications in user responses.
    - Prerequisite: User responses contain implicit information beyond literal statements.
    - `YES`: Agent fails to extract or use implied information that would aid task completion.
    - `NO`: The prerequisite IS met, but the conditions for `YES` are NOT satisfied.
    - `N/A`: The prerequisite is NOT met.
\end{lstlisting}

\subsection{Ask User Failure Modes in Failed Trials}\label{sec:fail-ask-fail-trials}

Table~\ref{tab:fail_with_questions} filters Table~\ref{tab:trial_with_questions} to trials where the task failed despite the agent asking questions. Failure rates rise across nearly every category---under-clarification increases from 26\% to 48\% on average, and question targeting errors from 41\% to 58\%---suggesting that question quality is a key bottleneck for recovering baseline performance.

\begin{table*}[h]
\centering
\caption{High-Level Ask User Failure Modes for Failed Trials with Questions}
\label{tab:fail_with_questions}
\resizebox{\textwidth}{!}{%
\begin{tabular}{@{}lrccccccc@{}}
\toprule
\textbf{Model} & \textbf{N} & \shortstack{\textbf{Question}\\\textbf{Quality}} & \shortstack{\textbf{Question}\\\textbf{Targeting}} & \shortstack{\textbf{Information}\\\textbf{Integration}} & \shortstack{\textbf{Over-}\\\textbf{Clarification}} & \shortstack{\textbf{Under-}\\\textbf{Clarification}} & \shortstack{\textbf{Timing \&}\\\textbf{Strategy}} & \shortstack{\textbf{Response}\\\textbf{Misinterp.}} \\
\midrule
Claude Opus 4.5   & 50  & 0.62 & 0.68 & 0.12 & 0.30 & 0.54 & 0.06 & 0.26 \\
Claude Sonnet 4.5 & 53  & 0.81 & 0.60 & 0.06 & 0.26 & 0.45 & 0.02 & 0.26 \\
Gemini 3 Flash    & 32  & 0.50 & 0.56 & 0.41 & 0.16 & 0.53 & 0.00 & 0.13 \\
Gemini 3 Pro      & 40  & 0.63 & 0.53 & 0.35 & 0.15 & 0.50 & 0.00 & 0.23 \\
GPT-5.2           & 134 & 0.98 & 0.60 & 0.22 & 0.46 & 0.36 & 0.05 & 0.37 \\
\bottomrule
\end{tabular}%
}
\end{table*}

\subsection{Sub-Taxonomy Ask User Failures for GPT-5.2}\label{sec:sub_taxonomy_failed_qns}

\begin{figure}[h]
\begin{center}
\centerline{\includegraphics[width=\columnwidth]{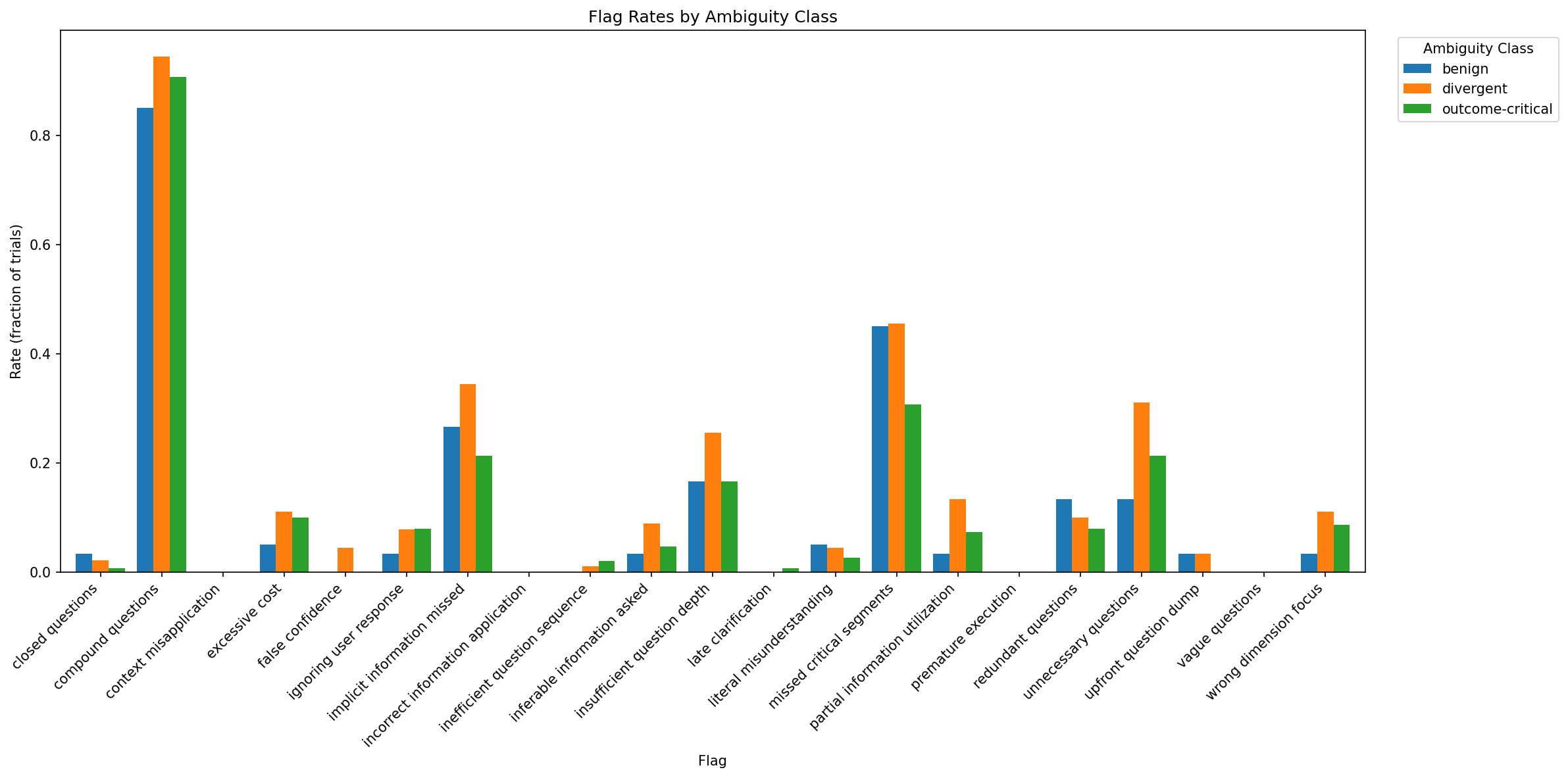}}
\caption{\textbf{Full Taxonomy of ask\_user Failure Modes.} Using the judge prompt in Section \ref{sec:ask-user-judge}, we measure the frequency of trials flagged with each failure mode across the full taxonomy. Compound questions and missed critical segments dominate across ambiguity classes.}
\label{fig:gpt_5_2_full_taxonomy}
\end{center}
\vskip -0.1in
\end{figure}

Figure~\ref{fig:gpt_5_2_full_taxonomy} breaks down ask\_user failure modes for GPT-5.2 by ambiguity class using the full sub-taxonomy from \cref{sec:ask-user-judge}. Compound questions and missed critical segments dominate across all classes, with outcome-critical variants showing the highest under-clarification rates.

\section{User Persona Details} \label{user_persona_appendix}
\begin{figure}[h]
\begin{center}
\centerline{\includegraphics[width=\columnwidth]{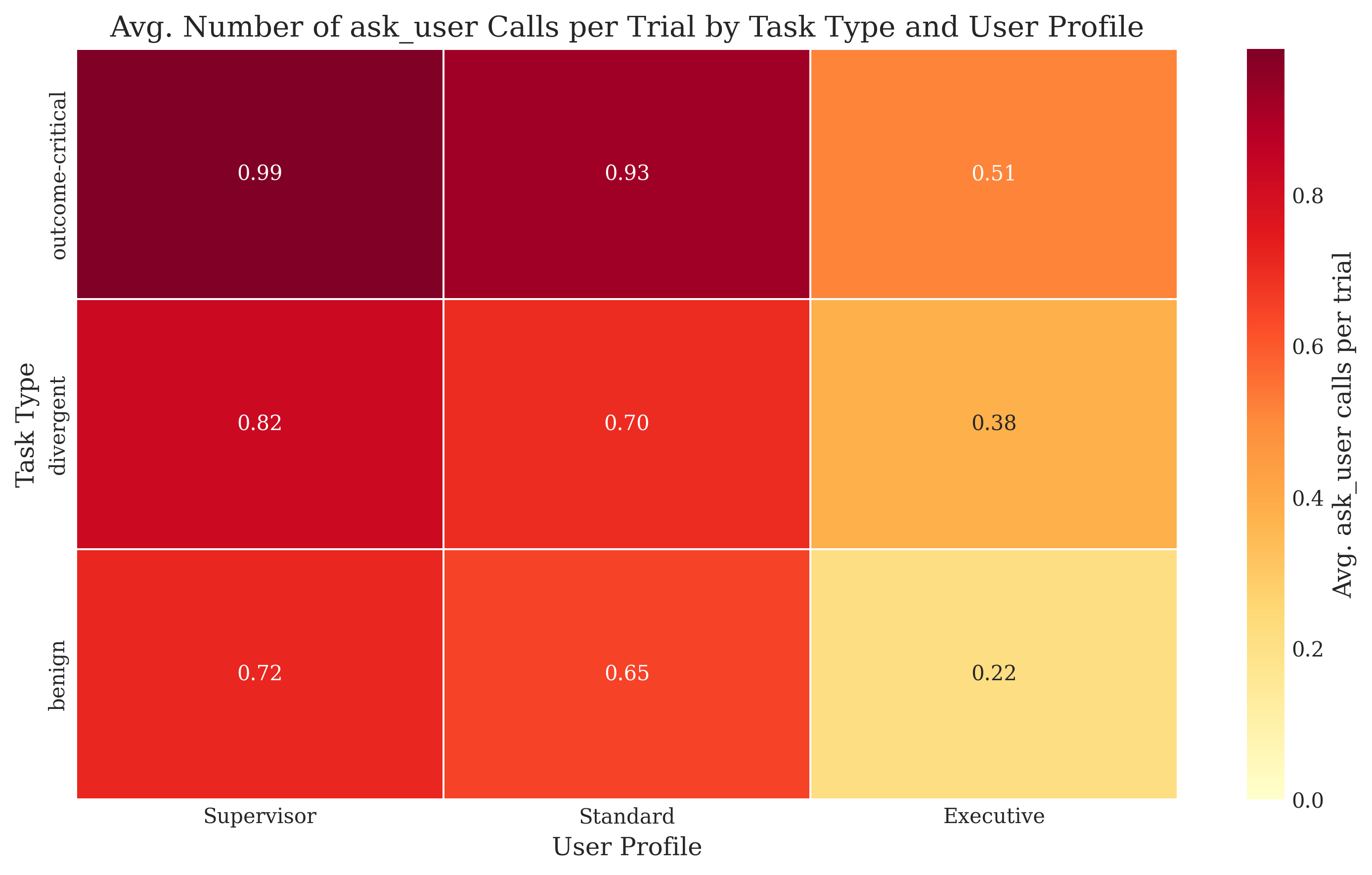}}
\caption{\textbf{Avg. number of ask\_user calls per trial} split by the \lhaw{} task type. Shows a natural gradient of persona (e.g. user cost) x value of information (e.g. implied blocking ambiguity). All tasks run on the \mcpatlas{} underspecified dataset using Claude Sonnet-4.5 as the agent.}
\label{fig:user_persona_heatmap}
\end{center}
\vskip -0.1in
\end{figure}

Figure~\ref{fig:user_persona_heatmap} shows average ask\_user calls per trial on \mcpatlas{} (Sonnet-4.5), broken down by ambiguity class and user persona. A natural gradient emerges from outcome-critical tasks under the Supervisor persona (0.99 calls/trial) to benign tasks under the Executive persona (0.22 calls/trial), indicating that agents modulate clarification frequency based on both task difficulty and perceived user availability.

\end{document}